\documentclass[10pt,twocolumn,letterpaper]{article}

\usepackage{iccv}
\usepackage{times}
\usepackage{epsfig}
\usepackage{graphicx}
\usepackage{amsmath}
\usepackage{amssymb}

\usepackage{soul}
\usepackage{subcaption}
\usepackage{caption}
\usepackage{tabularx}
\def\nbR{\ensuremath{\mathrm{I\! R}}}

\usepackage{multirow}
\usepackage{makecell}
\usepackage{comment}
\usepackage{pifont}
\usepackage{xcolor}
\newcommand{\cmark}{\ding{51}}%
\newcommand{\xmark}{\ding{55}}%

\newcommand{\datasetname}{HowToVQA69M}
\newcommand{\smalldatasetname}{iVQA}

\newcommand{\vqat}{VQA-T}
\newcommand{\qat}{QA-T}

\usepackage[pagebackref=true,breaklinks=true,letterpaper=true,colorlinks,bookmarks=false]{hyperref}

\iccvfinalcopy 

\def\iccvPaperID{4169} 

\begin{document}

\title{Just Ask: Learning to Answer Questions from Millions of Narrated Videos}

\author{Antoine Yang$^{1,2}$, Antoine Miech$^{1,2,+}$, Josef Sivic$^{3}$,
Ivan Laptev$^{1,2}$, Cordelia Schmid$^{1,2}$
\smallskip\\ 
\small{$^1$Inria Paris \quad $^2$D\'{e}partement d'informatique de l'ENS, CNRS, PSL Research University \quad $^3$CIIRC CTU Prague \quad $^+$Now at DeepMind}
\\
\small{\url{https://antoyang.github.io/just-ask.html}}
}
\maketitle

\begin{abstract}
Recent methods for visual question answering rely on
large-scale annotated datasets. 
Manual annotation of questions and answers for videos, however, is tedious, expensive and prevents scalability. 
In this work, we propose to avoid manual annotation and generate a large-scale training dataset for video question answering making use of automatic cross-modal supervision. 
We leverage a question generation transformer trained on text data and use it to generate question-answer pairs from transcribed video narrations. 
Given narrated videos, we then automatically generate the HowToVQA69M dataset with 69M video-question-answer triplets. 
To handle the open vocabulary of diverse answers in this dataset, we propose a training procedure based on a contrastive loss between a video-question multi-modal transformer and an answer transformer.  
We introduce the zero-shot VideoQA task and show excellent results, in particular for rare answers. 
Furthermore, we demonstrate our method to significantly outperform the state of the art on MSRVTT-QA, MSVD-QA, ActivityNet-QA and How2QA.
Finally, for a detailed evaluation we introduce \smalldatasetname{}, a new VideoQA dataset with reduced language biases and high-quality redundant manual annotations. 
\end{abstract}

\footnotetext[3]{Czech Institute of Informatics, Robotics and Cybernetics at the Czech Technical University in Prague.}

\vspace{-0.5cm}
\section{Introduction}\label{sec:intro}
Answering questions about videos requires a detailed understanding of the visual content and its association with the natural language. 
Indeed, given the large diversity of questions, methods for Video Question Answering (VideoQA) should reason about  scenes, objects and human actions as well as their complex temporal interactions. 

\begin{figure}[t]
\centering
\includegraphics[width=1.\linewidth]{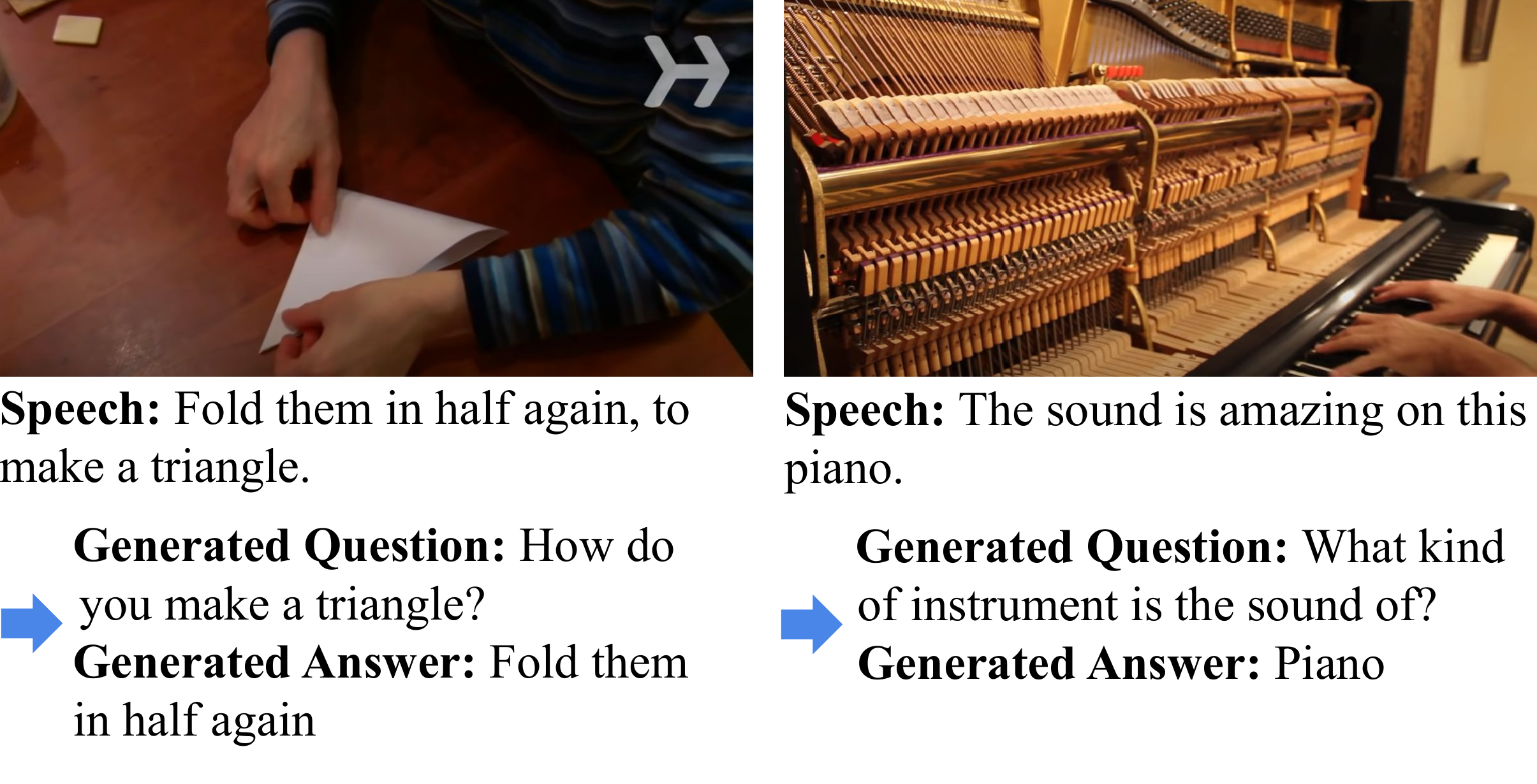}
\vspace{-0.8cm}
\caption{\small Given videos with transcribed narration, we leverage language models and cross-modal supervision to obtain large-scale VideoQA data. Above are two examples from our dataset.}
\label{fig:teasersqa}
\vspace{-0.5cm}
\end{figure}

\begin{figure}[t]
\centering
\includegraphics[width=\linewidth]{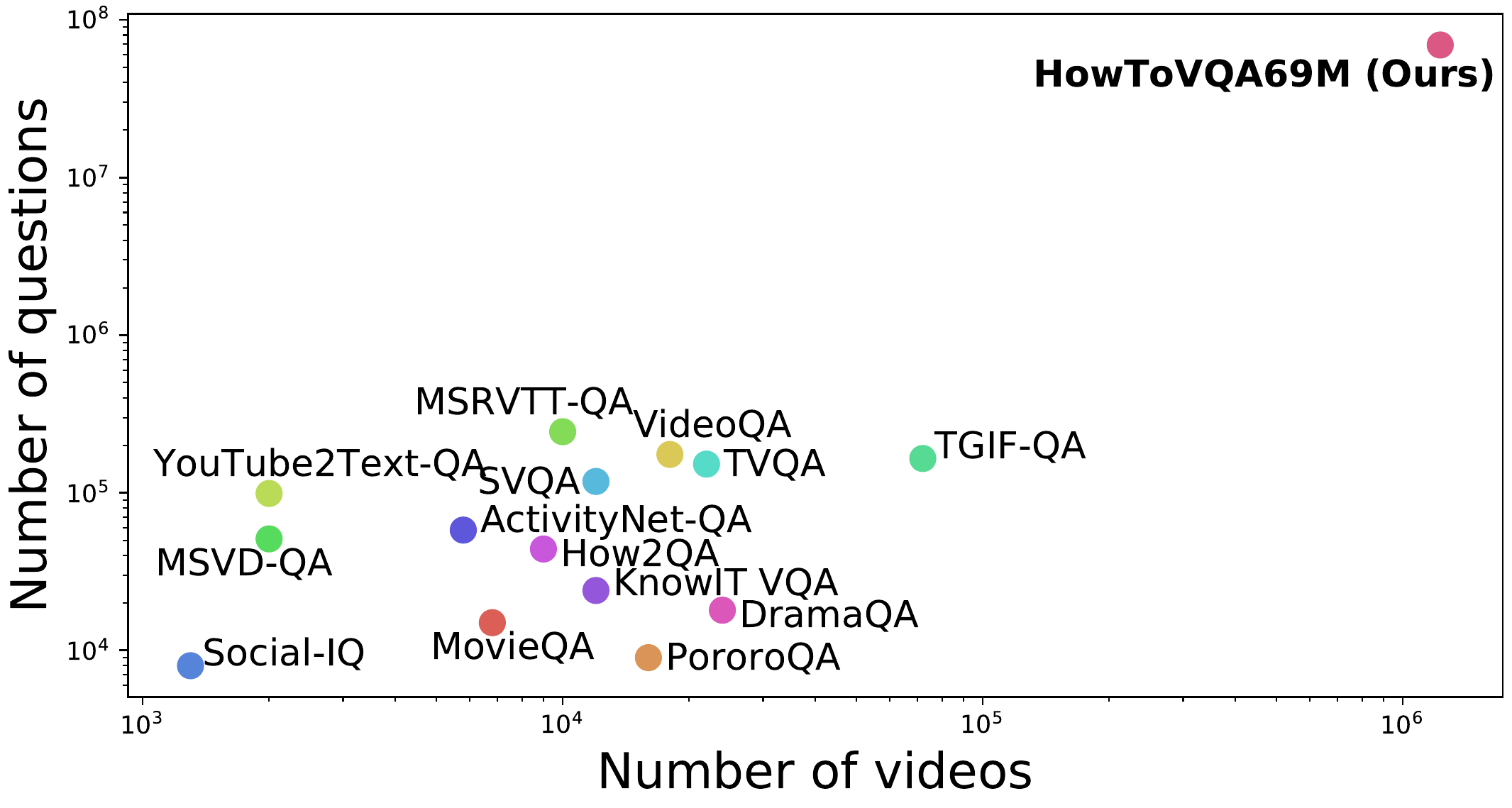}
\vspace{-0.7cm}
\caption{\small Comparison of our proposed large-scale \datasetname{} dataset with existing VideoQA datasets.}
\vspace{-0.7cm}
\label{fig:datasetcomparison}
\end{figure}

Current approaches to VideoQA  rely on deep fully-supervised models trained on manually annotated datasets with question and answer pairs~\cite{fan2019heterogeneous, huang2020location, jiang2020divide, jiang2020reasoning, le2020hierarchical, lei2021less, li2019beyond}.
Collecting and annotating VideoQA datasets, however, is cumbersome, time consuming, expensive and therefore not scalable. 
As a result, current VideoQA datasets are relatively small (see Figure~\ref{fig:datasetcomparison}). 
This limitation hinders the progress in the field as state-of-the-art VideoQA models often require a large amount of training data.

\begin{figure*}[t]
\centering
\includegraphics[width=\linewidth]{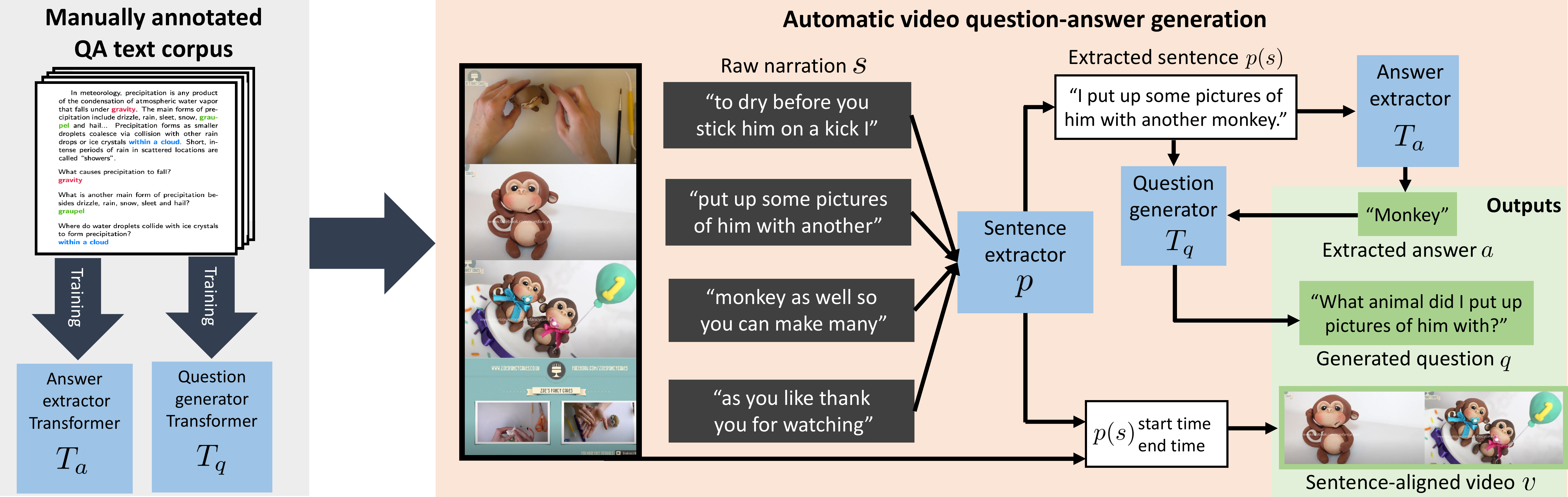}
\vspace{-0.7cm}
\caption{\small {\bf Our automatic approach for large-scale generation of video-question-answer triplets from narrated (subtitled) videos.} First, at the language-only training phase (left), the transformer-based answer extractor $T_a$ and question generator $T_q$ are trained~\cite{raffel2020exploring} on a manually annotated text-only question-answer corpus. Then video-question-answer triplets are automatically generated from narrated videos (right). Individual sentences are extracted from the ASR-transcribed narration using a punctuator $p$. Each extracted sentence is analyzed with an answer extractor $T_a$ and a question generator $T_q$ to produce answer $a$ and question $q$. The timestamps of the narration are used to obtain a video clip $v$ temporarily aligned to the extracted sentence to form the output video-question-answer triplet~$(v, q, a)$.}
\vspace{-0.6cm}
\label{fig:data_generation}
\end{figure*}

In this work, we address the scale issue with a new approach for automatically generating a VideoQA dataset, see Figure~\ref{fig:teasersqa} for examples. 
The idea is to leverage cross-modal supervision together with text-only tools for question generation and to automatically annotate VideoQA from a \emph{large amount of readily-available narrated videos}.
Inspired by the recent progress in language generation using transformer-based language models~\cite{brown2020language}, we leverage transformers trained on a question-answering text corpus to generate a diverse set of non-scripted questions and corresponding open-vocabulary answers from text. 
By applying these transformers to speech transcripts of narrated videos from the large-scale HowTo100M dataset~\cite{miech19howto100m}, we create \datasetname{}, an open-ended VideoQA dataset with 69 million video-question-answer triplets and a diverse set of more than 16M unique answers (see Figure~\ref{fig:data_generation}). 
As shown in Figure~\ref{fig:datasetcomparison}, our \datasetname{} is two orders of magnitude larger compared to prior VideoQA datasets.

Given the limited diversity of existing datasets, current methods typically reduce video question answering to a classification problem, where frequent answers are assigned to unique classes. Typically, up to 5K unique possible answers are considered. 
Such an approach, however, does not scale to the open vocabulary of 16M different answers in our dataset. 
To address this problem and to enable video question answering with highly diverse questions and answers, we introduce a training procedure based on contrastive learning  between a video-question multi-modal  transformer and an answer transformer that can handle free-form answers. This bypasses the need to define a discrete set of answer classes.

The goal of our work is to advance truly open-ended and generic solutions to VideoQA. To evaluate generalization,  we propose a new zero-shot VideoQA task 
where we prohibit any manual supervision of visual data during training.
Our VideoQA model, trained on~\datasetname{}, demonstrates excellent zero-shot results on multiple existing datasets, especially for rare answers. 
Moreover, when finetuned on target datasets, our model significantly outperforms the state of the art on 
MSRVTT-QA~\cite{xu2017video}, MSVD-QA~\cite{xu2017video} ActivityNet-QA~\cite{yu2019activitynet}, and How2QA~\cite{li2020hero}.

Initial experiments showed that existing benchmarks for open-ended VideoQA~\cite{xu2017video, yu2019activitynet} contain a language bias~\cite{goyal2017making}, i.e., their questions can often be answered without looking at the video. 
To better evaluate the impact of visual information in VideoQA, we introduce a new open-ended VideoQA dataset (iVQA) with manually collected questions and answers, where we exclude questions that could be answered without watching the video. 
Moreover, to account for multiple possible answers, iVQA contains five independently collected answers for each question.

In summary, our work proposes the following three contributions: 
\vspace{-.2cm}
\begin{itemize}
    \item[\textit{(i)}]
We introduce an approach to automatically generate a large-scale VideoQA dataset, \datasetname. Relying on cross-modal supervision, we use transformers trained on an existing text-only question-answering corpus and generate video-question-answer triplets from videos and transcribed narrations.
\vspace{-.2cm}
    \item[\textit{(ii)}]
We train a VideoQA model on \datasetname{} with contrastive learning between a multi-modal video-question transformer and an answer transformer.  We show the efficiency of our model in the new zero-shot VideoQA task and outperform the state of the art in four existing VideoQA benchmarks: MSRVTT-QA, MSVD-QA, ActivityNet-QA and How2QA. 
\vspace{-.2cm}
    \item[\textit{(iii)}]
Finally, we  introduce a new manually annotated open-ended VideoQA benchmark \smalldatasetname{} that excludes non-visual questions and contains multiple possible answers for each question.
\vspace{-.2cm}
\end{itemize}

 \noindent
 Code, datasets and trained models are available at \cite{justaskwebpage}. 

\section{Related Work}\label{sec:background}
\noindent \textbf{Visual Question Answering (VQA).}
VQA is typically tackled by classifying the  image-question (or video-question) representation into a fixed vocabulary of answers.
Various approaches to combine spatial image representations and sequential question representations have been proposed~\cite{anderson2018bottomup, ben2017mutan, fukui2016multimodal, lu2016hierarchical, xiong2016dynamic, xu2016ask, yang2016stacked}. More specifically to the video domain (VideoQA), spatio-temporal video representations in terms of motion and appearance have been used in~\cite{fan2019heterogeneous, gao2018motion, huang2020location, jang2017tgif, jiang2020divide, jiang2020reasoning, le2020hierarchical, le2020neural, lei2021less, li2019beyond, xu2017video, xue2018better, zha2019spatiotemporal, zhuang2020multichannel}. 

Methods above are limited to pre-defined vocabularies of answers and are difficult to apply outside of specific datasets.
To address this problem, Hu \etal~\cite{hu2018learning} propose a joint embedding where image-question representations can be matched with free-form answers.
Our VideoQA model follows this idea, but instead of relying on manually annotated datasets of limited scale, we train it on a large-scale VideoQA dataset that we automatically generate.
In contrast to some previous works using additional video features such as subtitles~\cite{chadha2020iperceive, kim2020dense, kim2020modality, lei2018tvqa, lei2019tvqa+, li2020hero, tapaswi16movieqa, winterbottom2020modality,yang2020bert}, our video representation is exclusively based on visual information, as we focus on the visual understanding of videos.

To evaluate the generalization of VQA models, Teney and Hengel~\cite{teney2016zero} define zero-shot VQA by answering previously unseen questions, which is a related but less challenging task compared to the zero-shot VQA task we propose in Section~\ref{sec:zeroshot}. Vatashsky and Ullman~\cite{vatashsky2020vqa} address VQA using COCO image annotations \cite{lin14coco}, while our zero-shot model is trained with no manual annotations. Our proposed zero-shot VQA task is analogous to zero-shot video retrieval~\cite{miech20endtoend} or zero-shot action recognition~\cite{radford2021learning}.

Visual question generation (VQG) has been introduced in \cite{mostafazadeh2016generating}. The methods in~\cite{li2018visual} and \cite{shah2019cycle} propose to jointly learn VQG and VQA to improve the image VQA task. However, these works do not generate questions to obtain additional training data, but use visual data annotation for question generation as an additional loss. 

\noindent \textbf{VideoQA datasets.}
Manually collecting and annotating video-question-answer triplets is cumbersome, costly and difficult to scale.
As result, current VideoQA datasets~\cite{castro2020lifeqa, choi2020dramaqa, colas2019tutorialvqa,fan2019egovqa,garcia2020knowit,jang2017tgif,kim2017deepstory, lei2018tvqa,li2020hero,mun2017marioqa, song2018explore, tapaswi16movieqa,xu2017video,ye2017video,yu2019activitynet, zadeh2019social, zeng2017leveraging} are limited in size, as the largest, TGIF-QA~\cite{jang2017tgif}, contains only 72K annotated clips (see Figure~\ref{fig:datasetcomparison} for more details).
To address this issue, several works have explored leveraging manually annotated video descriptions~\cite{jang2017tgif, wang2020long, xu2017video, zeng2017leveraging, zhao2020open, zhao2017video, zhao2018open} for automatic generation of VideoQA datasets, using rule-based~\cite{heilman2010good, ren2015exploring} approaches. 

Instead, we propose to use video narrations that are available at large-scale with no manual supervision. Moreover, rule-based generation requires the manual creation of rules by experts which is expensive, and has also been recently outperformed by neural question generation~\cite{du2017learning, yao2018teaching, zhou2017neural} as used in our approach.

\noindent \textbf{Large-scale pretraining for vision and language.}
Several recent methods~\cite{alberti2019fusion, chen2019uniter, desai2020virtex, huang2020pixel, li2019unicodervl, li2019visualbert, li2020oscar, lu2019vilbert, lu202012, su2019vl, tan2019lxmert, zhou2020unified} pretrain multi-modal vision-language representations, such as transformers, using datasets with image captions, e.g., COCO~\cite{chen2015microsoft}, Conceptual Captions~\cite{sharma2018conceptual} and Visual Genome~\cite{visualgenome}.
These methods are often optimized using generic objectives such as masked language losses and losses for text-image matching and image caption generation.
In our work, we pretrain models using large amounts of narrated videos. In contrast to task-agnostic pretraining in the previous work, we show the benefits of task-specific pretraining for our target VideoQA task.

\noindent \textbf{Learning from narrated videos.} 
In this work, we exploit noisy correlations between videos and narrations in unlabeled instructional videos from the recent HowTo100M dataset~\cite{miech19howto100m}.
Methods using such readily-available data have shown significant improvements on several tasks including video retrieval, action localization, action recognition and video captioning  \cite{gabeur2020multi, luo2020univilm,miech20endtoend,miech19howto100m,sun2019contrastive,sun2019videobert,zhu2020actbert}, sometimes outperforming fully-supervised baselines.
Some recent works use narrated videos for VideoQA.
Amrani \etal~\cite{amrani2020noise} propose a text-video pretraining approach and finetune for VideoQA.
Li \etal~\cite{li2020hero} propose HERO, a pretraining approach restricted to multiple-choice VideoQA, for which question and answer are treated as a single text stream.
Seo \etal~\cite{seo2020look} propose a pretraining approach based on next utterance prediction and finetune for VideoQA. 
Differently to these methods with task-agnostic pretraining, we propose a pretraining approach specifically dedicated for VideoQA using automatically generated question and answer pairs from narrated videos, and show in Section~\ref{sec:experiments} the superiority of our approach.

\section{Large-scale generation of VideoQA data}\label{sec:generation}
\begin{figure*}[t]
\begin{center}
\includegraphics[width=1.\linewidth]{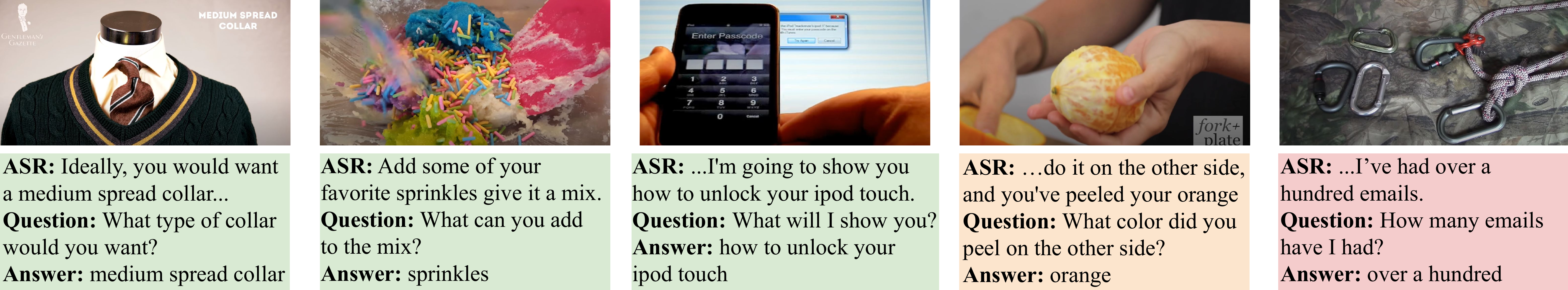}
\end{center}
\vspace{-0.6cm}
\caption{\small Examples of video-questions-answer triplets generated from narrated videos in our \datasetname{} dataset. {\color{green}The green color} indicates relevant examples, {\color{orange}the orange color} (penultimate example) indicates a failure of the question-answer generation, and {\color{red}the red color} (last example) indicates that the generated question-answer is unrelated to the visual content.}
\vspace{-0.6cm}
\label{fig:\datasetname{}}
\end{figure*}

This section presents our approach to generate a large-scale VideoQA dataset from videos and transcribed narrations describing the content of the videos. 
Section~\ref{sec:qgen} presents our proposed generation procedures. Section~\ref{sec:\datasetname{}}, then, describes the resulting \datasetname{} dataset.

\subsection{Generating video-question-answer triplets}\label{sec:qgen}
We tackle the task of generating video-question-answer triplets from a large-scale instructional video dataset with transcribed spoken narration~\cite{miech19howto100m}. 
This is a challenging task because of transcription errors and lack of punctuation. We also wish to obtain highly diverse data. To address these issues, we propose to leverage powerful language models trained on text data. Our approach is illustrated in Figure~\ref{fig:data_generation} and details are given next. 

We first present details about the generation procedure. Let $s$ be the transcribed speech data obtained with automatic speech recognition (ASR). First, we use a recurrent neural network $p$, to infer punctuation in the transcribed speech data.
We denote the punctuated transcript as $p(s)$. 
We extract video clips $v$ temporally aligned with the inferred sentences $p(s)$ using the ASR timestamps. 
We found that the generation works significantly better when applied to sentences rather than the original sentence fragments from the HowTo100M dataset, see Table~\ref{table:manual}.
Second, for each sentence, we apply a transformer $T_a$, to extract a set of potential answers:  $a = T_a(p(s))$.
Third, we use another transformer $T_q$ to generate a question given each transcript sentence and each extracted answer such that: $q = T_q(a, p(s))$. 
The output is a set of video-question-answer triplets $(v, q, a)$.

We now explain details about the language models and their training procedure. For ASR, we follow \cite{miech19howto100m} and use the readily-available ASR data provided by YouTube. For punctuation $p$, we use the BRNN model from~\cite{tilk2016} and the weights available at~\cite{punct} trained on IWSLT2011~\cite{federico2012iwslt}. 
For $T_a$ and $T_q$, we use the transformer-based T5-small and T5-base models~\cite{raffel2020exploring}, respectively. 
We follow~\cite{alberti2019synthetic, chan2019recurrent, lopez2020transformer} and use the weights available at~\cite{qgen} trained for answer span extraction and answer-aware question generation, respectively, on SQuADv1~\cite{rajpurkar2016squad}.
SQuADv1 is a text-only question-answering dataset consisting of questions for which the answer is a segment of text extracted from a paragraph.

\subsection{\datasetname{}: large-scale VideoQA dataset}\label{sec:\datasetname{}}
We have applied the previously described procedure to all 1.2M original videos from the HowTo100M dataset~\cite{miech19howto100m}.
The result is \datasetname{}, a dataset of 69,270,581 video clip, question and answer triplets $(v, q, a)$.
\datasetname{} is two orders of magnitude larger than any of the currently available VideoQA datasets (see Figure~\ref{fig:datasetcomparison}). 
On average, each original video results in 43 video clips, where each clip lasts 12.1 seconds and is associated to 1.2 question-answer pairs. 
Questions and answers contain 8.7 and 2.4 words on average respectively.
\datasetname{} is highly diverse and contains over 16M unique answers, where over 2M unique answers appear more than once and over 300K unique answers appear more than ten times. 
Examples of $(v, q, a)$ triplets from the \datasetname{} dataset are illustrated in Figure \ref{fig:\datasetname{}}. 

\vspace*{1mm}
\noindent \textbf{Manual evaluation of \datasetname{}.}\label{sec:eval}
As shown in Figure \ref{fig:\datasetname{}}, \datasetname{} annotations are noisy, which can be attributed to: (i) errors in speech transcription, (ii) speech not describing the video content, or (iii) errors in question-answer generation. 
We manually evaluated the quality of 100 randomly sampled $(v, q, a)$ triplets in \datasetname{}, collected 5 different annotations for each triplet to reduce variance, and reported results in Table~\ref{table:manual}. 
Among 100 triplets generated by our method we find 30 to be  correctly generated and matching well to the video content, 31 are incorrectly generated and 39 are correctly generated but unrelated to the video content. 
To demonstrate the influence of the different components of our automatic question-answer generation procedure, we compare it with (i) a variant of our approach that does not split transcribed narrations into sentences using a punctuator, and (ii) a rule-based approach~\cite{heilman2010good} for question-answer generation.
Table~\ref{table:manual} confirms the importance of punctuation and demonstrates the superior performance of our generation method compared to~\cite{heilman2010good}.
Inter-rater agreement statistics, and more details for the generated dataset are provided in Appendix \ref{sec:sqadata}.
Further comparison with~\cite{heilman2010good} is given in Section~\ref{sec:java}. 
We describe next how we use \datasetname{} to train our VideoQA model.

\begin{table}[t]
\begin{center}
\setlength\tabcolsep{2pt}
\resizebox{\linewidth}{!}{	
\begin{tabular}{ll|ccc}
Punctuation & Generation method & \makecell{ \small{Correct} \\ \small{Samples}} & \makecell{ \small{QA Generation} \\ \small{Failure}} & \makecell{ \small{QA unrelated} \\ \small{to video}} \\ 
\hline
\cmark & Heilman \etal~\cite{heilman2010good} & 17 & 54 & 29 \\
\xmark & Ours & 23 & 49 & 28 \\
\cmark & Ours & \textbf{30} & 31 & 39 \\
\end{tabular}
}
\end{center}
\vspace{-0.6cm}
\caption{\small Manual evaluation of our generation method (with and without punctuation) on a random sample of 100 examples compared with a rule-based question-answer generation of~\cite{heilman2010good}. Numbers are obtained with majority voting between 5 annotators.}
\vspace{-0.6cm}
\label{table:manual}
\end{table} 

\section{VideoQA model and training procedure}\label{sec:learning}
This section presents our VideoQA model in Section \ref{sec:model} and describes its training procedure in Section \ref{sec:training}. Figure~\ref{fig:videoqamodel} gives an overview of the model.

\subsection{VideoQA model}\label{sec:model}

\begin{figure}[t]
\centering
\includegraphics[width=0.95\linewidth]{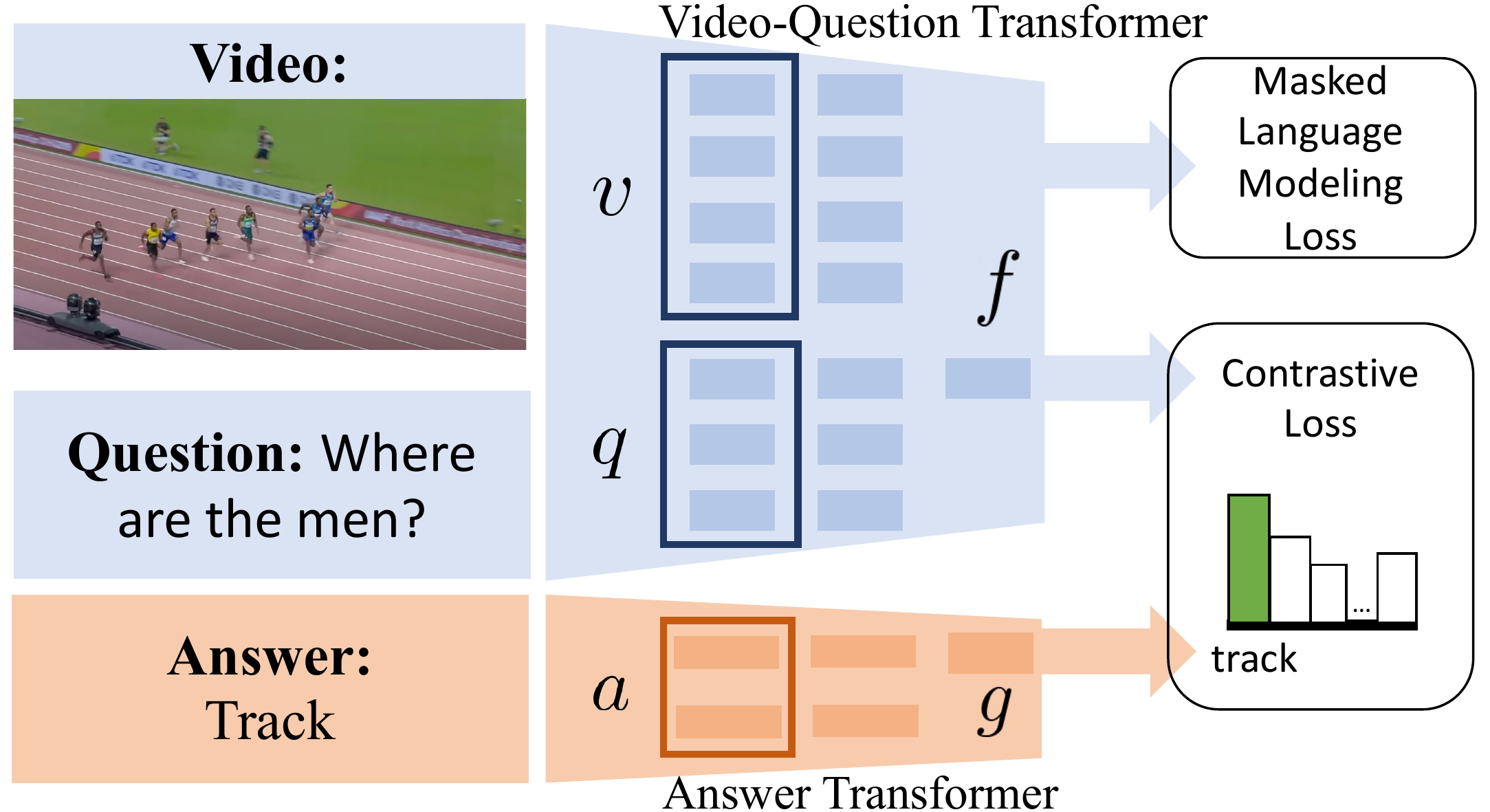}
\vspace{-0.3cm}
\caption{\small Overview of our VideoQA training architecture.}
\vspace{-0.6cm}
\label{fig:videoqamodel}
\end{figure}

As illustrated in Figure \ref{fig:videoqamodel}, our VideoQA model is composed of two branches: \textit{(i)} a video-question module $f$ based on a transformer~\cite{vaswani2017attention} and a mapping from the CLS token with a linear function.
It takes a pair of video $v$ and question $q$ as input, models the multi-modal temporal interactions between $v$ and $q$ and then outputs an embedding vector $f(v,q) \in \nbR^{d}$. \textit{(ii)} The second branch is a text encoder $g$ that embeds an answer $a$ as $g(a) \in \nbR^{d}$. 
We will denote our model as \textit{VQA-T}, standing for VideoQA-Transformer. 
Note that using the joint (video, question) and answer embeddings allows us to deal with a large open vocabulary of answers present in our new  \datasetname{} dataset as the model can measure similarity between the input video-question embedding and the embedding of any answer. This is in contrast to using a classification answer module~\cite{huang2020location, jiang2020divide, jiang2020reasoning, le2020hierarchical, zhuang2020multichannel} that can choose only from a fixed predefined vocabulary of answers. Our embedding can be also easily finetuned on the different downstream VideoQA datasets, which may contain new answers that have not been seen at training. In contrast, the classification answer module has to be retrained when the vocabulary of answers changes. 
Next, we give details of the language and video representations. Further details about the model are provided in Appendix \ref{sec:mmt}.

\vspace*{1mm}
\noindent \textbf{Word representation.}
The question and answer are separately tokenized with the WordPieces embedding~\cite{wu2016google} and fed to DistilBERT~\cite{sanh2019distilbert}. DistilBERT is a light version of BERT~\cite{bert18} pretrained in a self-supervised fashion on English Wikipedia and the Toronto Book Corpus \cite{zhu15aligning}.

\vspace*{1mm}
\noindent \textbf{Video representation.}
We use a frozen S3D~\cite{xie2018rethinking} pretrained on HowTo100M~\cite{miech19howto100m} using MIL-NCE~\cite{miech20endtoend}. This model is pretrained from scratch on HowTo100M only.

\subsection{Training procedure}\label{sec:training}
This section describes the training of our VideoQA model on the \datasetname{} dataset and its finetuning on downstream VideoQA datasets.

\vspace{-0.4cm}
\paragraph{Training on \datasetname{}.}
We wish to make a pair of video and question $(v,q)$ close to its correct answer $a$  measured by the dot product of their embeddings, $f(v,q)^\top g(a)$.
Conversely, the incorrect answers should be far, i.e., the dot product with their embeddings should be small. 
Formally, this can be done by maximizing the following contrastive objective:
\begin{equation}
\label{eq:objective}
\max_{f,g} \sum_{i=1}^n\log\left(\frac{e^{ f(v_i,q_i)^\top g(a_i)}}{e^{ f(v_i,q_i)^\top g(a_i)}+\sum\limits_{(v',q',a')\sim\mathcal{N}_i}e^{f(v',q')^\top g(a')}}\right),
\end{equation}
where $(v_i,q_i,a_i)$ represents a triplet of generated (video clip, question, answer) from \datasetname{}.
Given a specific positive triplet $(v_i,q_i,a_i)$, we construct the set $\mathcal{N}_i$ of negative triplets by concatenating incorrect answers $a_j$ within the training batch to the video-question pair $(v_i, q_i)$ as: $(v_i,q_i,a_j)$ with $a_j \neq a_i$. 
In particular, if the same negative answer $a_j$ is present multiple times in a batch, we only count it once.
We found that sampling the same negative answer multiple times leads to worse results (see Section~\ref{sec:ablations}), which we believe is due to different distributions of answers in the pretraining and downstream datasets. Removing duplicate negatives helps to mitigate this difference.

\vspace{-0.4cm}
\paragraph{Finetuning on downstream VideoQA datasets.}\label{sec:finetune}
We leverage the model pretrained on \datasetname{} and finetune it on a downstream VideoQA dataset that typically has a smaller vocabulary of answers $V$ (\eg $|V| \sim 4000$). 
To this end, we adapt the training objective in \eqref{eq:objective} by constructing the negative set $\mathcal{N}_i$ from {\em all} incorrect answers in $V$. Note that in such setting \eqref{eq:objective} becomes equivalent to optimizing the standard cross-entropy objective. In the specific case of multiple-choice VideoQA, the set of negatives $\mathcal{N}_i$ is the set of incorrect answers for each sample.

\vspace{-0.4cm}
\paragraph{Masked Language Modeling (MLM).}\label{sec:mlm} 
In addition to the contrastive loss~(\ref{eq:objective}) we apply the masking loss~\cite{bert18} to question tokens during both pretraining and finetuning. We found this to have a positive regularization effect when finetuning the DistilBERT weights (see Section~\ref{sec:ablations}).

\section{\smalldatasetname{}: new dataset for VideoQA evaluation}\label{sec:ivqa}
In this section we present our {\bf I}nstructional {\bf V}{\bf QA} dataset (iVQA). We start from a subset of HowTo100M videos and manually annotate video clips with questions and answers. 
We aim to (i) provide a well-defined evaluation by including five correct answer annotations per question and (ii)~avoid questions which can be answered without watching the video. 
The dataset is described below and more details are given in Appendix \ref{sec:ivqadata} and \ref{sec:bias}.

\vspace{-0.4cm}
\paragraph{Data Collection.}\label{sec:collection}
iVQA videos are obtained by randomly sampling 7-30 sec.\ video clips from the HowTo100M dataset~\cite{miech19howto100m}. 
We avoid overlap between datasets and make sure iVQA and \datasetname{} have no videos in common.
Each clip is manually annotated with one question and 5 answers on Amazon Mechanical Turk.
We ask workers to annotate questions about objects and scenes in the video and remove videos that could not be annotated. 
The correctness of annotations is manually verified by the authors. Moreover, we manually reduce the language bias by excluding questions that could be answered without watching the video.
To increase diversity, each question is answered by 5 different workers. The answers are restricted to 4 words and are complemented by a confidence level. 
Questions that receive multiple answers with low confidence are removed.

\begin{table*}[t]
\vspace{-0pt}
\begin{center}
\resizebox{.9\linewidth}{!}{
\begin{tabular}{lc|ccccccccc}
Method & Pretraining Data & 
\multicolumn{2}{c}{\smalldatasetname{}} & \multicolumn{2}{c}{MSRVTT-QA} & 
\multicolumn{2}{c}{MSVD-QA} & 
\multicolumn{2}{c}{ActivityNet-QA} & 
How2QA
\\ 
& & Top-1 & Top-10 & Top-1 & Top-10 & Top-1 & Top-10 & Top-1 & Top-10 & Top-1 \\
\hline
Random & $\emptyset$
& 0.09 & 0.9 & 0.02 & 0.2 & 0.05 & 0.5 & 0.05 & 0.5 & 25.0 \\
\qat{} & \datasetname{}
& 4.4 & 23.2 & 2.5 & 6.5 & 4.8 & 15.0 & 11.6 & 45.8 & 38.4 \\
\vqat{} & HowTo100M &
1.9 & 11.9 & 0.3 & 3.4 & 1.4 & 10.4 & 0.3 & 1.9 & 46.2 \\
\vqat{} (Ours) & \datasetname{}
& \textbf{12.2} & \textbf{43.3} & \textbf{2.9} & \textbf{8.8} & \textbf{7.5} & \textbf{22.4} & \textbf{12.2} & \textbf{46.5} & \textbf{51.1} \\
\end{tabular}}
\vspace{-0.3cm}
\caption{\small Comparison with baselines for zero-shot VideoQA. Top-1 and top-10 (for open-ended datasets) accuracy are reported.}
\vspace{-.8cm}
\label{table:zeroshot}
\end{center}
\end{table*}

\vspace{-0.4cm}
\paragraph{Statistical Analysis.}\label{sec:analysis}
\smalldatasetname{} contains 10,000 video clips with one question and five corresponding answers per clip. We split the dataset into 60\%/20\%/20\% train/validation/test subsets. 
On average, questions and answers contain 7.6 and 1.1 words respectively.
The average duration of video clips is 18.6 seconds.
The majority of questions have at least 2 annotators providing the same answer. Similarly to \cite{antol2015vqa}, this motivates us to define the following accuracy measure for a given answer $a$:
$acc(a)=\min(\frac{\#\textrm{ground truth answers =}\, a}{2},1).$
This metric assigns 100\% accuracy to answers confirmed by at least 2 annotators, 50\% accuracy to answers confirmed by only 1 annotator and 0\% otherwise. Note that this definition is specific to \emph{multiple} ground truth answers per question.

\section{Experiments}\label{sec:experiments}
\begin{figure}[t]
\centering
\includegraphics[width=1.\linewidth]{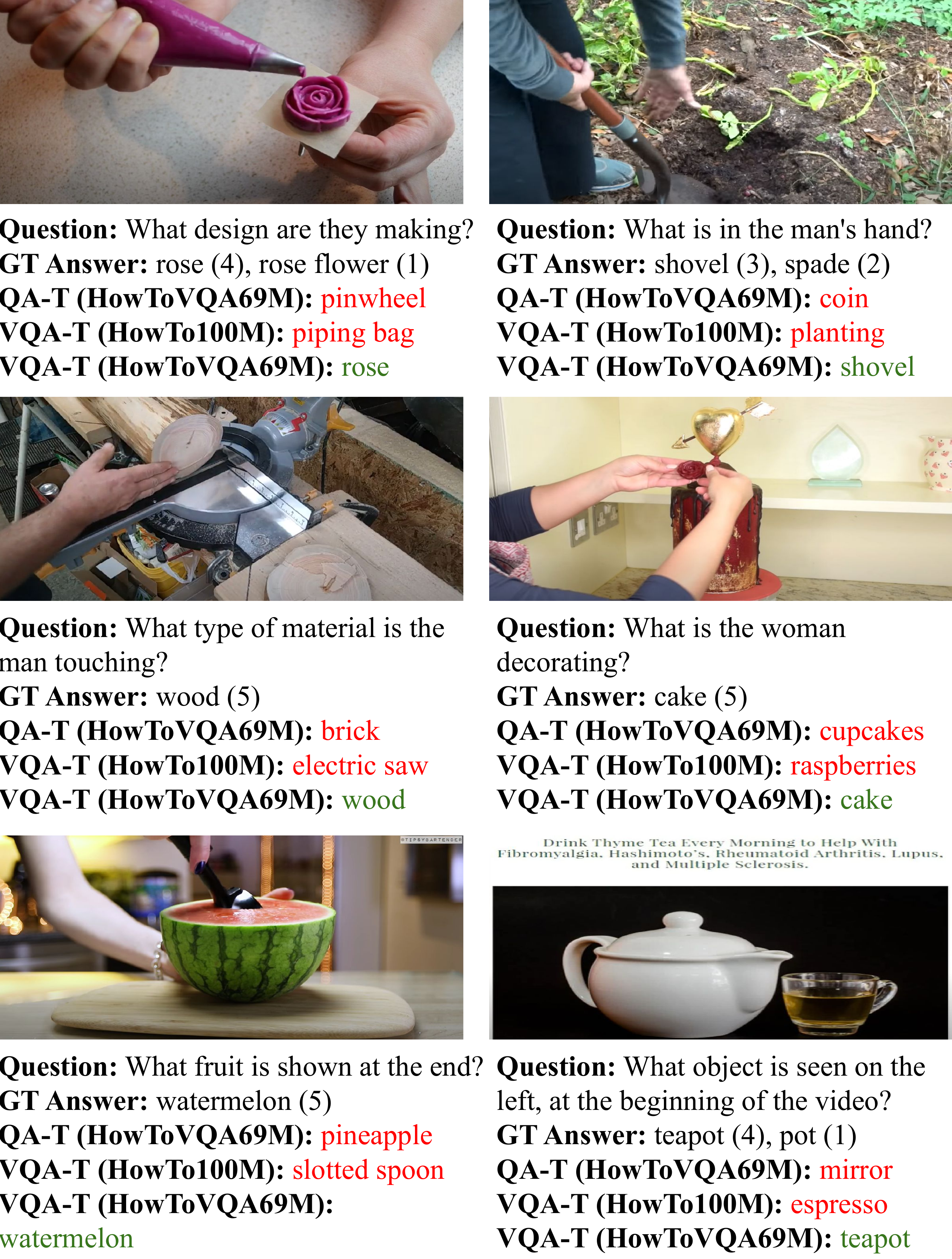}
\vspace{-0.5cm}
\caption{\small \textbf{Zero-shot VideoQA on \smalldatasetname{}}. The values next to the ground truth (GT) answers indicate the number of annotators that gave the answer.
}
\label{fig:zeroshot}
\vspace{-0.3cm}
\end{figure}

This section demonstrates the benefits of training using our generated \datasetname{} dataset and compares our method to the state of the art.
We first outline the used datasets, baseline methods and implementation details in Section \ref{sec:protocol}.
We then present results for the novel zero-shot VideoQA task in Section~\ref{sec:zeroshot}. The comparison to the state of the art in VideoQA and alternative training strategies is given in Section~\ref{sec:results}. Section~\ref{sec:rare} presents results for rare answers.
Finally, we compare our VideoQA generation approach to previous methods in Section~\ref{sec:java} and present ablation studies in Section~\ref{sec:ablations}.

\subsection{Evaluation Protocol}\label{sec:protocol}

\paragraph{Datasets.}\label{sec:datasets}
We use two datasets for training and five datasets for evaluation as described below. 
We follow previous evaluation protocols for open-ended settings~\cite{le2020hierarchical, yu2019activitynet} and use a fixed vocabulary of training answers.
Unless stated otherwise, we report top-1 test accuracy and use original splits for training, validation and test. 

For training we use our new \textbf{\datasetname{}} dataset introduced in Section~\ref{sec:\datasetname{}} with 90\% and 10\% videos in training and validation subsets. 
For comparison, we also train our model using a large-scale text-video dataset,~\textbf{HowTo100M}~\cite{miech19howto100m}, that contains videos with transcribed narrations but \emph{no video-question-answer} triplets.
Test and validation videos of downstream datasets are excluded from HowTo100M and \datasetname{}.

We evaluate results on four open-ended VideoQA downstream datasets: \textbf{MSRVTT-QA}~\cite{xu2017video}, \textbf{MSVD-QA}~\cite{xu2017video}, \textbf{ActivityNet-QA}~\cite{yu2019activitynet} and our new \textbf{\smalldatasetname{}} dataset (see Section~\ref{sec:ivqa}).
We also evaluate on a multiple-choice VideoQA dataset \textbf{How2QA}~\cite{li2020hero} where each question is associated with one correct and three incorrect answers. 

\vspace{-.4cm}
\paragraph{Baselines.}\label{sec:baselines}
To evaluate the contribution of the visual modality, we compare our \textit{\vqat{}} model with its language-only variant \textit{\qat{}}.
\textit{\qat{}} does not use video input, i.e.~we set the input $v$ of the video-question transformer to zero (see Figure~\ref{fig:videoqamodel}).
To evaluate our generated dataset, we also compare \textit{\vqat{}} trained on~\datasetname{} and on HowTo100M.  
Since HowTo100M has no $(v,q,a)$ triplets, we only train the $f$ branch of \textit{\vqat{}} on HowTo100M using the standard masking and cross-modal matching losses \cite{chen2019uniter, li2020hero, lu2019vilbert, sun2019videobert, zhu2020actbert}. 
In the zero-shot setting we evaluate \textit{\vqat{}} trained on HowTo100M by computing $f(v,[q,a])$ for concatenated pairs of questions and answers $[q,a]$.
During finetuning we also initialize the $g$ branch of \textit{\vqat{}} with parameters of the text encoding obtained from $f$ (see further details in Appendix \ref{sec:mmt}).

\vspace{-.4cm}
\paragraph{Implementation details.}\label{sec:details}
For the training on \datasetname{} we use the Adam optimizer and mini-batches with 4096 video clips sampled from 128 random videos. 
The optimization over 10 epochs lasts 2 days on 8 Tesla V100 GPUs. 
Further details are included in Appendix \ref{sec:detailsbis}.

\subsection{Zero-shot VideoQA}\label{sec:zeroshot}

In this section, we address the {\em zero-shot VideoQA} task 
where we prohibit any manual supervision of visual data during training.
We explore this setup to evaluate the generalization of \textit{\vqat{}} trained on \datasetname{} to unseen downstream datasets.
For consistency, we use the vocabulary of answers from downstream datasets during testing (see Section~\ref{sec:datasets}). 

Zero-shot results are presented in Table \ref{table:zeroshot}. 
We first observe that the use of visual cues by \textit{\vqat{}} outperforms \mbox{\textit{\qat{}}} when both models are trained on \datasetname{}. 
This demonstrates the importance of the cross-modality in \datasetname{} despite the VideoQA annotation being exclusively generated from text-only methods.
Since \datasetname{} has been generated using no manual annotation of visual data, our approach is scalable and can lead to further improvements by increasing the dataset size, as we discuss in Section~\ref{sec:ablations}. 

Training on \datasetname{} significantly outperforms the training on HowTo100M and the random baseline. 
This confirms the advantage of our \datasetname{} dataset for the VideoQA task over other generic text-video datasets that do not contain video-question-answer triplets.
We emphasize that our training does not use any information about target VideoQA datasets.
Qualitative results for zero-shot VideoQA are presented for our approach and compared with baselines in Figure~\ref{fig:zeroshot}. 
We observe that \textit{\qat{}} (trained on \datasetname{}) provides plausible but video-unrelated answers to the questions. 
Moreover, \textit{\vqat{}} (trained on HowTo100M) is able to associate visual content with related answers, but fails to have a complex multi-modal understanding.
Our \textit{\vqat{}} model trained on \datasetname{}, on the other hand, correctly understands questions and uses information in the video to provide correct answers, confirming results in Table~\ref{table:zeroshot}.

\begin{table}[t]
\begin{center}
\setlength\tabcolsep{1.5pt}
\resizebox{\linewidth}{!}{	
\begin{tabular}{c|ccccc}
Pretraining data & \smalldatasetname{} & \makecell{ \small{MSRVTT} \\ \small{QA}} & \makecell{\small{MSVD} \\ \small{QA}} & \makecell{\small{ActivityNet} \\ \small{QA}} & \small{How2QA} \\
\hline
$\emptyset$ & 23.0 & 39.6 & 41.2 & 36.8 & 80.8 \\
HowTo100M & 28.1 & 40.4 & 43.5 & 38.1 & 81.9 \\ 
\datasetname{} & \textbf{35.4} & \textbf{41.5} & \textbf{46.3} & \textbf{38.9} & \textbf{84.4} \\
\end{tabular}
}
\end{center}
\vspace{-0.6cm}
\caption{\small Benefits of pretraining our \textit{VQA-T} model on our new HowToVQA69M dataset (last row) compared to no pretraining (first row) or pretraining on HowTo100M (second row). In each case our \textit{VQA-T} model was then finetuned on the downstream VideoQA datasets. Top-1 accuracy is reported.}
\vspace{-0.6cm}
\label{table:baselines}
\end{table} 

\begin{table}[t]
\begin{center}
\setlength\tabcolsep{3pt}
\resizebox{\linewidth}{!}{	
\begin{tabular}{lc|ccc}
Method & Pretraining data & \small{MSRVTT-QA} & \small{MSVD-QA} \\
\hline
E-SA \cite{xu2017video} & & 29.3 & 27.6 \\ 
ST-TP \cite{jang2017tgif} & & 30.9 & 31.3 \\
AMU \cite{xu2017video} & & 32.5 & 32.0 \\
Co-mem \cite{gao2018motion} & & 32.0 & 31.7 \\ 
HME \cite{fan2019heterogeneous} & & 33.0 & 33.7 \\ 
LAGCN \cite{huang2020location} & & --- & 34.3 \\  
HGA \cite{jiang2020reasoning} & & 35.5 & 34.7 \\ 
QueST \cite{jiang2020divide} & & 34.6 & 36.1 \\ 
HCRN \cite{le2020hierarchical} & & 35.6 & 36.1 \\ 
ClipBERT \cite{lei2021less} & \makecell{ \small{COCO \cite{chen2015microsoft}+} \\ \small{Visual Genome \cite{visualgenome}} }  & 37.4 & --- \\  
SSML \cite{amrani2020noise} & HowTo100M & 35.1 & 35.1 \\
CoMVT \cite{seo2020look} & HowTo100M & 39.5 & 42.6 \\ 
\hline
\vqat{} & $\emptyset$ & 39.6 & 41.2 \\ 
\vqat{} & \datasetname{} & \textbf{41.5} & \textbf{46.3} \\ 
\end{tabular}
}
\end{center}
\vspace{-0.6cm}
\caption{\small Comparison with state of the art on MSRVTT-QA and MSVD-QA (top-1 accuracy).}
\vspace{-0.4cm}
\label{table:sotamvtmvd}
\end{table} 

\begin{table}[t]
\setlength\tabcolsep{2.5pt}
\centering
\resizebox{.95\linewidth}{!}{	
\begin{tabular}{lc|cc}
& Pretraining data & \makecell{ \small{ActivityNet} \\ \small{QA} } & \small{How2QA} \\
\hline
E-SA \cite{yu2019activitynet} & & 31.8 & --- \\
MAR-VQA \cite{zhuang2020multichannel} & & 34.6 & --- \\
HERO \cite{li2020hero} & \makecell{ \small{HowTo100M +} \\ \small{TV Dataset} } & --- & 74.1 \\
CoMVT \cite{seo2020look} & HowTo100M & 38.8 & 82.3 \\
\hline
\vqat{} & $\emptyset$ & 36.8 & 80.8 \\
\vqat{} & \datasetname{} & \textbf{38.9} & \textbf{84.4} \\
\end{tabular}
}
\vspace{-0.3cm}
\caption{\small Comparison with state of the art on ActivityNet-QA and the public val set of How2QA (top-1 accuracy).}
\vspace{-0.2cm}
\label{table:sotaacthow2qa}
\end{table}

\subsection{Benefits of \datasetname{} pretraining}\label{sec:results}

This section evaluates the effect of \textit{\vqat{}} pretraining in combination with finetuning on target datasets.
As shown in Table~\ref{table:baselines}, pretraining on \datasetname{} provides consistent and significant improvements for all datasets when compared to pretraining on HowTo100M and no pretraining.
In particular, we observe the largest improvement for our new \smalldatasetname{} dataset which comes from the same domain as \datasetname{}. Hence, the automatic generation of training data for other domains using our method can lead to further improvements on other datasets. 

We compare our pretrained model to the state-of-the-art in VideoQA in Tables~\ref{table:sotamvtmvd}-\ref{table:sotaacthow2qa}.
Notably, \textit{\vqat{}} pretrained on \datasetname{} outperforms previous methods on all tested datasets. In particular, our method improves over the recent CoMVT approach~\cite{seo2020look} that has been pretrained on HowTo100M.
These strong results show the importance of our proposed \datasetname{} dataset.

\begin{table}[t]
\centering
\resizebox{\linewidth}{!}{	
\begin{tabular}{cc|cccc}
Pretraining data & Finetuning & Q1 & Q2 & Q3 & Q4\\
\hline
$\emptyset$ & \cmark & 38.4 & 16.7 & 5.9 & 2.6 \\
HowTo100M
& \cmark & 46.7 & 22.0 & 8.6 & 3.6 \\[1ex]
\multirow{2}{*}{\datasetname{}} & \xmark
& 9.0 & 8.0 & 9.5 & 7.7 \\
& \cmark & \textbf{47.9} & \textbf{28.1} & \textbf{15.6} & \textbf{8.5} \\
\end{tabular}
}
\vspace{-0.3cm}
\caption{\small 
Results of our \textit{\vqat{}} model with different training strategies, on subsets of \smalldatasetname{} corresponding to four quartiles with Q1 and Q4 corresponding to samples with most frequent and least frequent answers, respectively. 
}
\vspace{-0.3cm}
\label{table:rare}
\end{table}

\subsection{Results for rare answers}\label{sec:rare}
Training on downstream VideoQA datasets typically leads to particularly large improvements for questions with most frequent answers. 
As shown in Table \ref{table:rare}, our approach brings significant improvements both for common and rare answers compared to models trained from scratch or pretrained on HowTo100M.
Interestingly, for the most rare answers in \smalldatasetname{} (Q3 and Q4) our model without finetuning (zero-shot mode) outperforms finetuned models that have not been pretrained on \datasetname{}.
We make similar observations for rare answers in other datasets and report corresponding results in Appendix \ref{sec:rarebis}. 
We conclude that VideoQA specific pretraining on additional large-scale, diverse data helps improve generalization of VideoQA models. 

\begin{table}[t]
\vspace{-0pt}
\setlength\tabcolsep{1.5pt}
\begin{center}
\resizebox{\linewidth}{!}{
\begin{tabular}{l|ccc|ccc}
\makecell{Generation \\ Method} & \multicolumn{3}{c}{Zero-shot} & \multicolumn{3}{c}{\small{Finetune}} \\
& \smalldatasetname{} & \makecell{\small{ActivityNet} \\ \small{QA}} & \small{How2QA} & \smalldatasetname{} & \makecell{\small{ActivityNet} \\ \small{QA}} & \small{How2QA} \\
\hline
\cite{heilman2010good} & 7.4 & 1.1 & 41.7 & 31.4 & 38.5 & 83.0 \\
Ours & \textbf{12.2} & \textbf{12.2} & \textbf{51.1} & \textbf{35.4} & \textbf{38.9} & \textbf{84.4} \\
\end{tabular}
}
\vspace{-0.3cm}
\caption{\small{Comparison of our question-answer generation approach with Heilman \etal~\cite{heilman2010good}, evaluated by downstream performance of the model trained on the generated VideoQA data.}} 
\label{table:java}
\end{center}
\vspace{-0.8cm}
\end{table}

\subsection{Comparison of VideoQA generation methods}\label{sec:java}
In this section, we compare our question-answer generation approach to Heilman \etal~\cite{heilman2010good}, that was notably used in \cite{xu2017video, zeng2017leveraging, zhao2020open, zhao2017video,  zhao2018open} to generate VideoQA data from video descriptions.
We run the method of~\cite{heilman2010good} on sentences extracted from HowTo100M, apply our pretraining method on the generated data and show results in Table~\ref{table:java}. 
Note that we do not choose MSRVTT-QA and MSVD-QA as downstream datasets for this comparison because their evaluation sets were automatically generated using Heilman \etal~\cite{heilman2010good}. 
We find that our generation method leads to significantly better performance both in zero-shot and finetuning settings. 
We also provide a qualitative comparison in Appendix \ref{sec:sqadata}, further demonstrating the benefit of our transformer-based question-answer generation approach compared to previous methods.
We also show the benefit of our generated \datasetname{} dataset by comparing our results to cross-dataset transfer using existing VideoQA datasets in Appendix \ref{sec:transfer}. 

\subsection{Ablation studies}\label{sec:ablations}

\noindent \textbf{Pretraining losses.} As shown in Table~\ref{table:loss}, removing duplicate negative answers in our contrastive loss, as discussed in Section~\ref{sec:training}, is beneficial notably in the zero-shot setting.
Moreover, adding the MLM loss at pretraining improves the downstream results for both zero-shot and finetuning when used in combination with our contrastive learning strategy.
These results motivate our proposed pretraining approach.

\begin{table}[t]
\vspace{-0pt}
\setlength\tabcolsep{2pt}
\begin{center}
\resizebox{\linewidth}{!}{
\begin{tabular}{cc|cc|cc}
MLM & \makecell{ \small{Sampling without} \\ \small{answer repetition} } &
\multicolumn{2}{c}{Zero-shot} & \multicolumn{2}{c}{Finetune} \\
& & \smalldatasetname{} & \small{MSVD-QA} & 
\smalldatasetname{} & \small{MSVD-QA} \\
\hline
\xmark & \xmark & 11.1 & 6.1 & 34.7 & 45.6 \\
\xmark & \cmark & 12.1 & 7.0 & 34.3 & 45.0 \\
\cmark & \xmark & 10.9 & 6.4 & 34.3 & 45.1 \\
\cmark & \cmark & \textbf{12.2} & \textbf{7.5} & \textbf{35.4} & \textbf{46.3} \\
\end{tabular}
}
\vspace{-0.3cm}
\caption{\small{Effect of MLM loss and our negative sampling strategy on \datasetname{} training.}}
\label{table:loss}
\end{center}
\vspace{-0.7cm}
\end{table}

\noindent \textbf{Importance of scale.}
Results of our method after pretraining on different fractions of \datasetname{} are shown in Table~\ref{table:scale}. 
We construct these subsets such that larger subsets include the smaller ones. 
These results suggest that the scale is an important factor and that we can expect further improvements with additional pretraining data, both in the zero-shot and finetuning settings.

\begin{table}[t]
\vspace{-0pt}
\setlength\tabcolsep{3pt}
\begin{center}
\resizebox{\linewidth}{!}{
\begin{tabular}{l|cc|cc}
Pretraining data size &
\multicolumn{2}{c}{Zero-shot} & \multicolumn{2}{c}{Finetune} \\
& \smalldatasetname{} & \small{MSVD-QA} & 
\smalldatasetname{} & \small{MSVD-QA} \\
\hline
0\% & --- & --- & 23.0 & 41.2 \\
1\% & 4.5 & 3.6 & 24.2 & 42.8 \\
10\% & 9.1 & 6.2 & 29.2 & 44.4 \\
20\% & 9.5 & 6.8 & 31.3 & 44.8 \\
50\% & 11.3 & 7.3 & 32.8 & 45.5 \\
100\% & \textbf{12.2} & \textbf{7.5} & \textbf{35.4} & \textbf{46.3}  \\
\end{tabular}
}
\vspace{-0.3cm}
\caption{\small{Effect of the training size of \datasetname{}.}}
\label{table:scale}
\end{center}
\vspace{-1cm}
\end{table}

\section{Conclusion}\label{sec:conclusion}
We propose a novel and scalable approach for training VideoQA models without manually annotated visual data.
We automatically generate \datasetname{} -- a large-scale VideoQA training dataset generated from narrated videos with readily-available speech transcripts, significantly exceeding existing datasets by size and diversity.
We demonstrate several benefits of pretraining on \datasetname{}. We are the first to 
demonstrate zero-shot VideoQA results without the use of any manually annotated images or videos. 
Furthermore,  finetuning our \datasetname{} pretrained model on downstream tasks outperforms the state of the art on MSRVTT-QA, MSVD-QA, ActivityNet-QA and How2QA.
We further validate our approach on a new \smalldatasetname{} benchmark we manually collect.

\mbox{}\vspace{-.2cm}\\
\noindent 
{\footnotesize{
{\textbf{Acknowledgements.} This work was granted access to the HPC resources of IDRIS under the allocation 2020-101267 made by GENCI. The work was funded by a Google gift,  the French government under management of Agence Nationale de la Recherche as part of the "Investissements d'avenir" program, reference ANR-19-P3IA-0001 (PRAIRIE 3IA Institute), the Louis Vuitton ENS Chair on Artificial Intelligence, the European Regional Development Fund under project IMPACT (reg.\ no.\ CZ.02.1.01/0.0/0.0/15 003/0000468) and A. Miech's Google PhD fellowship.
We thank P.-L. Guhur and M. Tapaswi for advice on using Amazon Mechanical Turk, E. Berthier, Q. Le Lidec and E. Chane-Sane for the manual evaluation of generated VideoQA data, and I. Rocco for proofreading.}
}}

{\small
\bibliographystyle{ieee_fullname}
\bibliography{egbib}
}

\clearpage \newpage
\appendix

\section*{Appendix}
In this Appendix, we start by giving additional  analysis and examples of our proposed \datasetname{} dataset in Section \ref{sec:sqadata}.
We, then,  provide additional architecture details for our VideoQA model in Section \ref{sec:mmt}.
Next, we present additional statistics and details of the collection procedure for our manually collected \smalldatasetname{} evaluation benchmark in Section \ref{sec:ivqadata}.
We describe additional implementation details in Section \ref{sec:detailsbis} and present experiments including cross-dataset transfer, results per answer quartile and per question type in Section \ref{sec:expbis}.

\section{Analysis of \datasetname{} dataset}
\label{sec:sqadata}

\begin{figure}[t]
\centering
\begin{subfigure}{.5\textwidth}
\includegraphics[width=\linewidth]{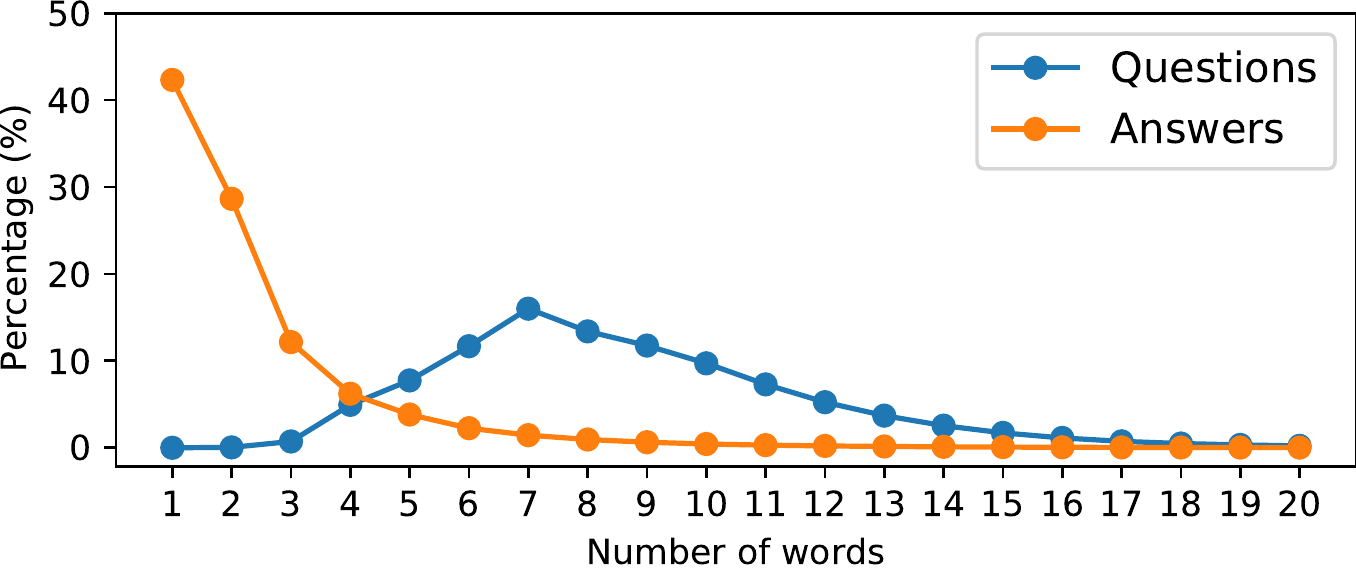}
\caption{Question and answer length}
\end{subfigure}
\begin{subfigure}{.5\textwidth}
\includegraphics[width=\linewidth]{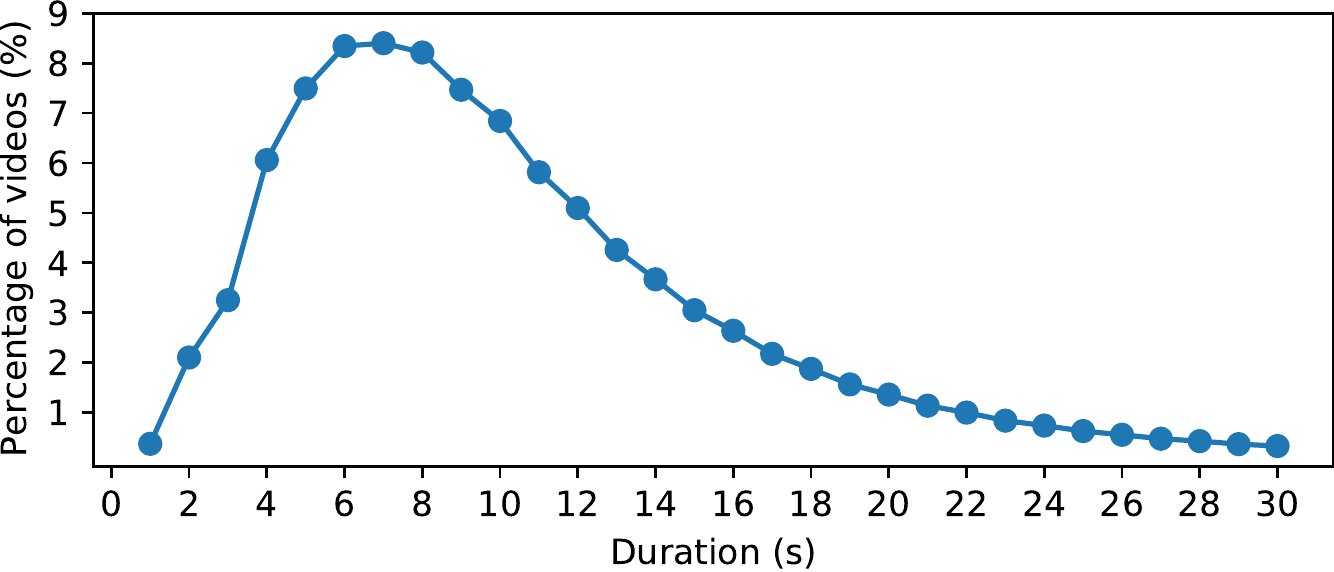}
\caption{Clip duration}
\end{subfigure}
\vspace{-0.3cm}
\caption{{\bf Statistics of the \datasetname{} dataset.} (a)~Distribution of length of questions and answers. (b)~Distribution of video clip duration in seconds.}
\label{fig:sqa_length}
\vspace{-0.3cm}
\end{figure}

Figure \ref{fig:sqa_length} shows the statistics of the \datasetname{} dataset in terms of the question length, answer length and video clip duration. 
Overall, \datasetname{} contains longer answers than downstream open-ended VideoQA datasets like MSRVTT-QA, MSVD-QA or ActivityNet-QA.
The distribution of clip duration has a peak at around seven seconds with a long tail of longer clips.  These statistics demonstrate the diversity of our \datasetname{} dataset, both in terms of videos and answers.

\begin{figure*}[t]
\centering
\begin{subfigure}{0.8\textwidth}
\includegraphics[width=\linewidth]{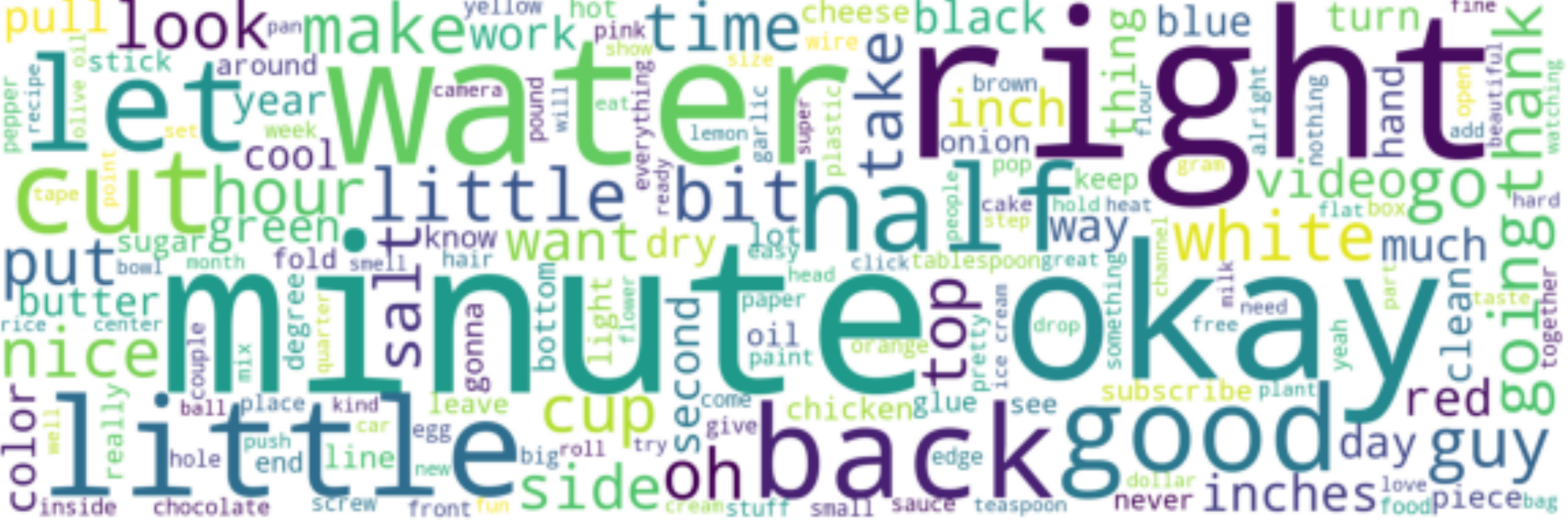}
\caption{Answers}
\end{subfigure}
\begin{subfigure}{0.8\textwidth}
\includegraphics[width=\linewidth]{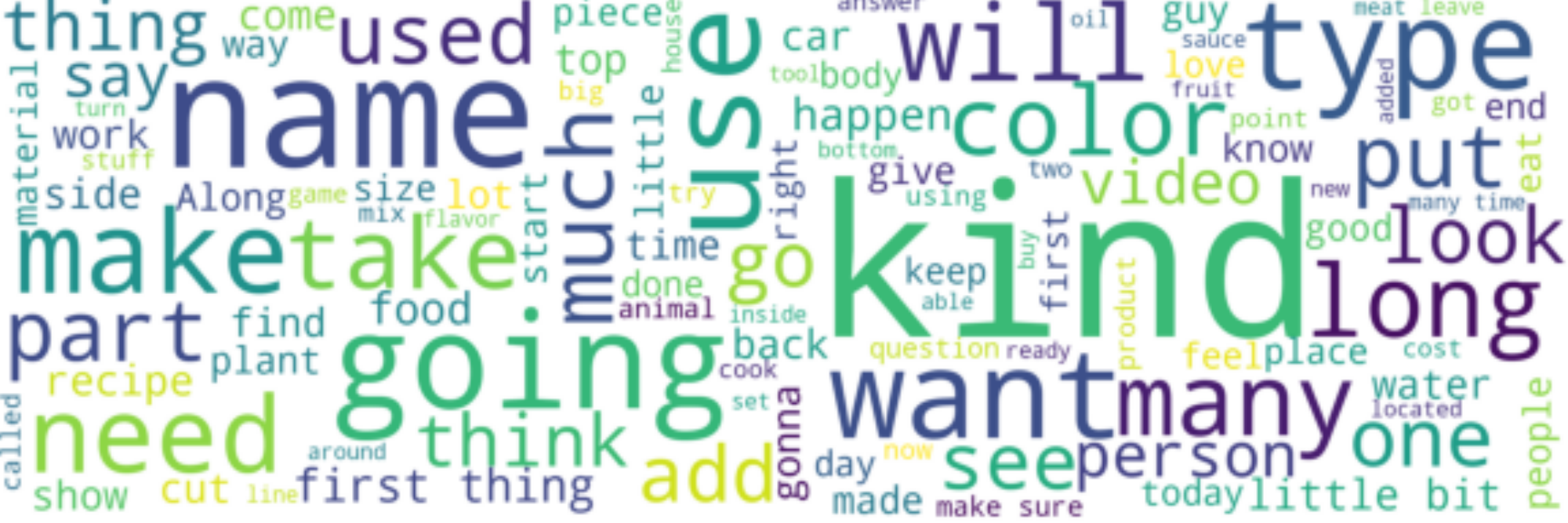}
\caption{Questions}
\end{subfigure}
\caption{Word clouds extracted from the \datasetname{} dataset showing its diverse vocabulary and the words characteristic to speech such as \textit{okay}, \textit{right}, or \textit{ok}.}
\label{fig:sqa_words}
\end{figure*}

\begin{figure*}[t]
\centering
\includegraphics[width=\linewidth]{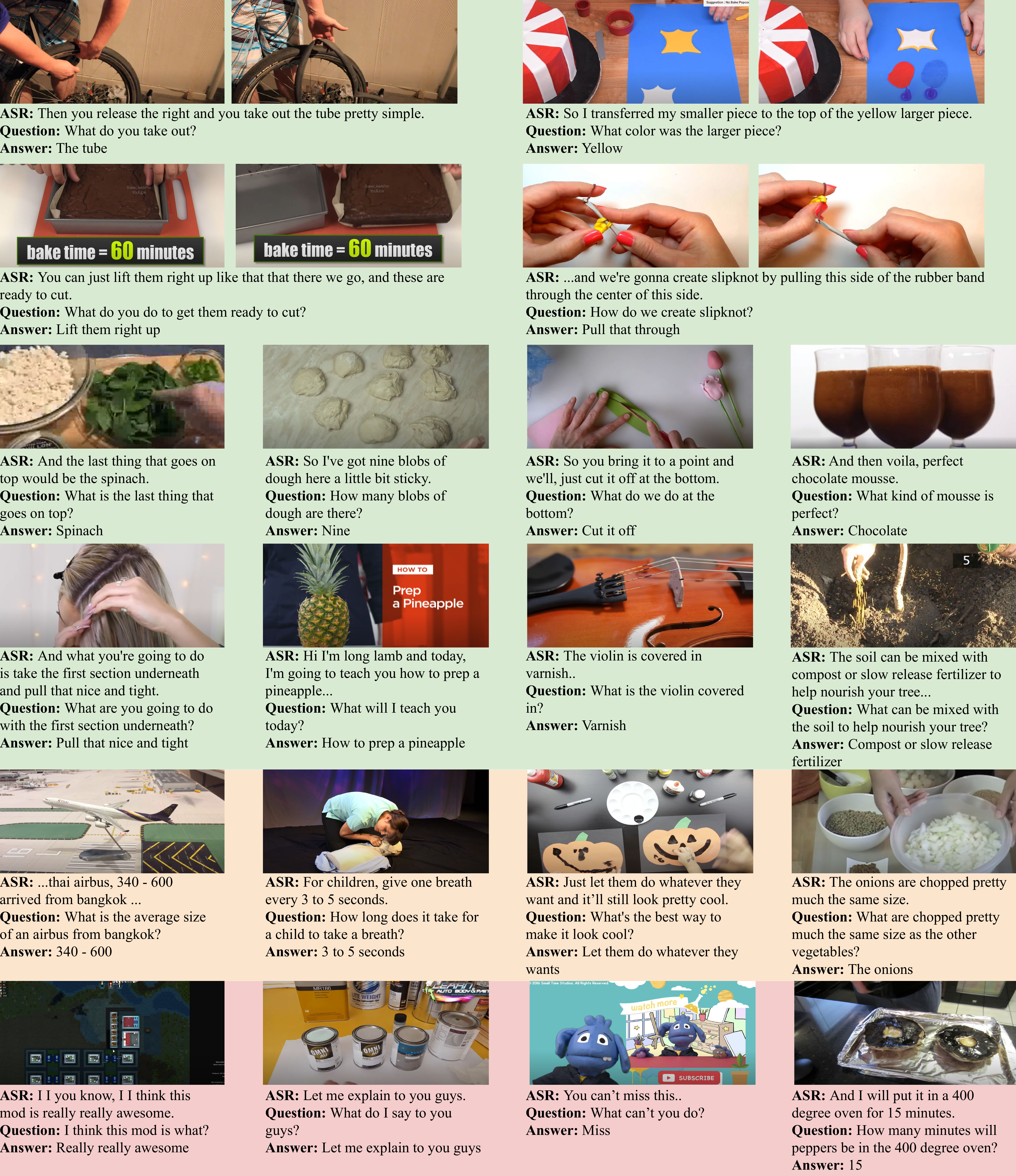}
\caption{Additional examples of videos, questions and answers from our automatically generated \datasetname{} dataset. These examples illustrate the large data diversity in \datasetname{}. {\color{green}The green color} indicates relevant examples, {\color{orange}the orange color} (penultimate row) indicates a failure of the question-answer generation, and {\color{red}the red color} (last row) indicates that the generated question-answer is unrelated to the visual content.}
\label{fig:sqa_add}
\end{figure*}

Word clouds\footnote{To generate the word clouds, we used \url{https://github.com/amueller/word_cloud}.} for questions and answers in \datasetname{} are shown in Figure~\ref{fig:sqa_words} and illustrate the diverse vocabulary in \datasetname{} as well as the presence of speech-related words such as as \textit{okay}, \textit{right}, \textit{oh}. 
In Figure~\ref{fig:sqa_add} we illustrate the diversity and the noise in the automatically obtained annotations in the \datasetname{} dataset.

\begin{figure*}[t]
\centering
\includegraphics[width=1.\linewidth]{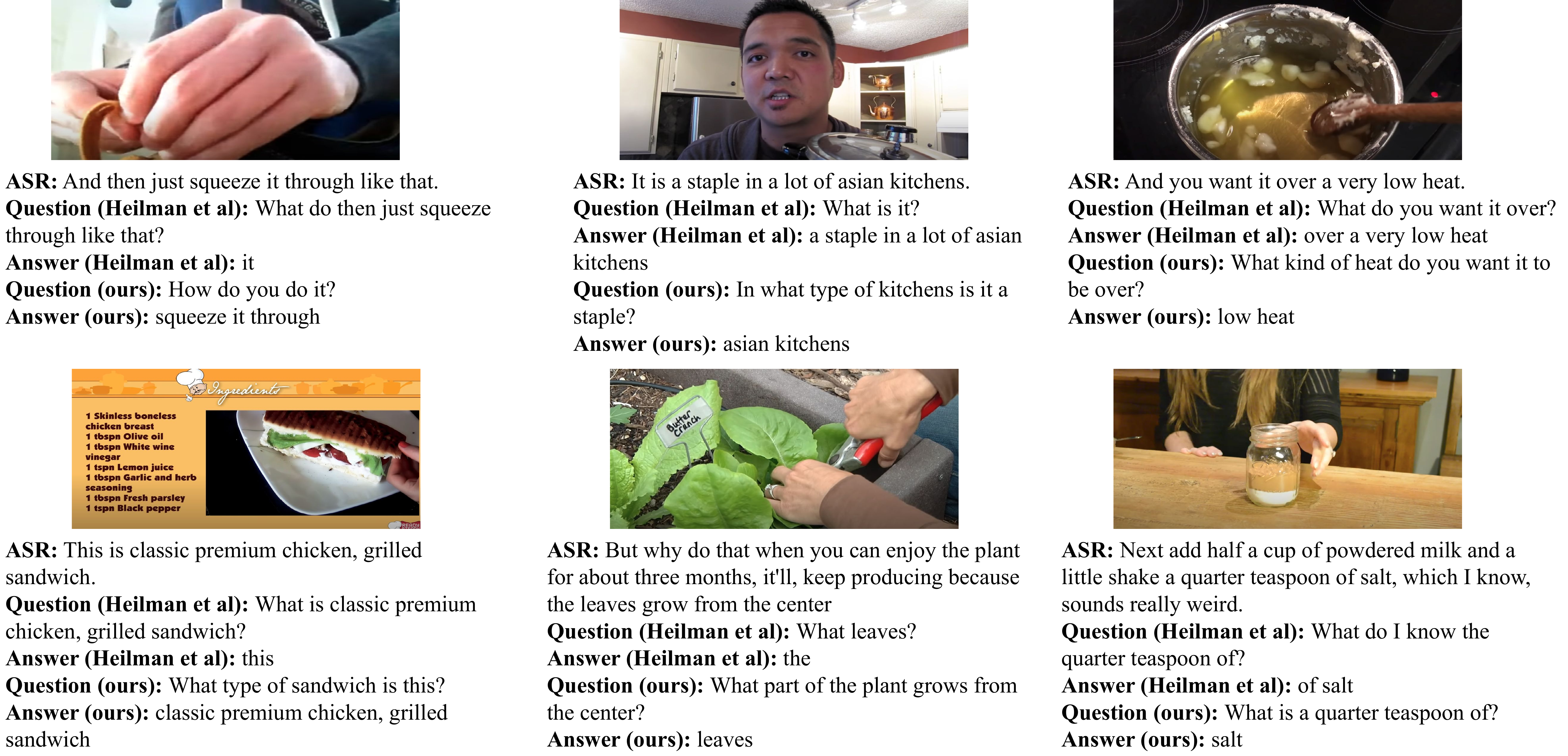}
\caption{Qualitative examples of video-question-answer triplets generated with our trained language models compared to Heilman \etal~\cite{heilman2010good}, illustrating the higher quality and diversity of triplets obtained with our generation method.}
\label{fig:java}
\end{figure*}

We show quantitative comparisons of our question-answer generation models with~\cite{heilman2010good} in Section \ref{sec:java}, and supplement it here with a qualitative comparison shown in Figure~\ref{fig:java}. 
We found that compared to~\cite{heilman2010good} our generation method provides higher quality as well as higher diversity of question-answer pairs when applied to the uncurated sentences extracted from speech in narrated videos.

In Section \ref{sec:\datasetname{}} we present a manual evaluation of the quality of the automatically generated video-question-answer triplets for our method and two other baselines. We complement this analysis here with inter-rater agreement statistics.
For the 300 generated video-question-answer triplets (100 for each generation method), 94 were in an agreement of all 5 annotators, 198 in an agreement of at least 4 annotators, and 299 in an agreement of at least 3 annotators. 
This high agreement of annotators demonstrates the reliability of the results in Table \ref{table:manual}. 

We further manually classify the 100 video-question-answer triplets obtained with our method by the question type (``Attribute", ``Object", ``Action", ``Counting", ``Place", ``People", or ``Other"), evaluate the quality of generated triplets for different question types and report results in Table~\ref{table:manualsplit}.
Out of the 6 most common categories, we observe that questions related to ``Action" lead to the best annotations, ``Counting" questions lead to the highest number of QAs unrelated to the video content, and questions related to ``Place" lead to the highest number of QA generation errors. 
Qualitatively, we found that actions are often depicted in the video, while counted quantities (\eg time, weight, length) mentioned in the speech are hard to guess from the video only. 

\begin{table}[t]
\begin{center}
\setlength\tabcolsep{1.5pt}
\resizebox{0.75\linewidth}{!}{%
\begin{tabular}{ll|ccc}
\makecell{ \small{Question} \\ \small{Type}} & Total &
\makecell{ \small{Correct} \\ \small{Samples (\%)}} & \makecell{ \small{QA Generation} \\ \small{Failure (\%)}} & \makecell{ \small{QA unrelated} \\ \small{to video (\%)}} \\ 
\hline
Attribute & 25 & 28 & 32 & 40 \\
Object & 17 & 41 & 24 & 35 \\
Action & 16 & \textbf{69} & 19 & 13 \\
Counting & 13 & 23 & 15 & \textbf{62} \\
Place & 7 & 0 & \textbf{86} & 14 \\
People & 7 & 0 & 43 & 57 \\
Other & 15 & 13 & 27 & 60 \\
\end{tabular}%
}
\end{center}
\vspace{-0.4cm}
\caption{\small Manual evaluation of our video-question-answer generation method on 100 randomly chosen generated examples split by question type. Results are obtained by majority voting among 5 annotators.}
\vspace{-0.6cm}
\label{table:manualsplit}
\end{table} 

\begin{figure*}[t]
\centering
\includegraphics[width=.8\linewidth]{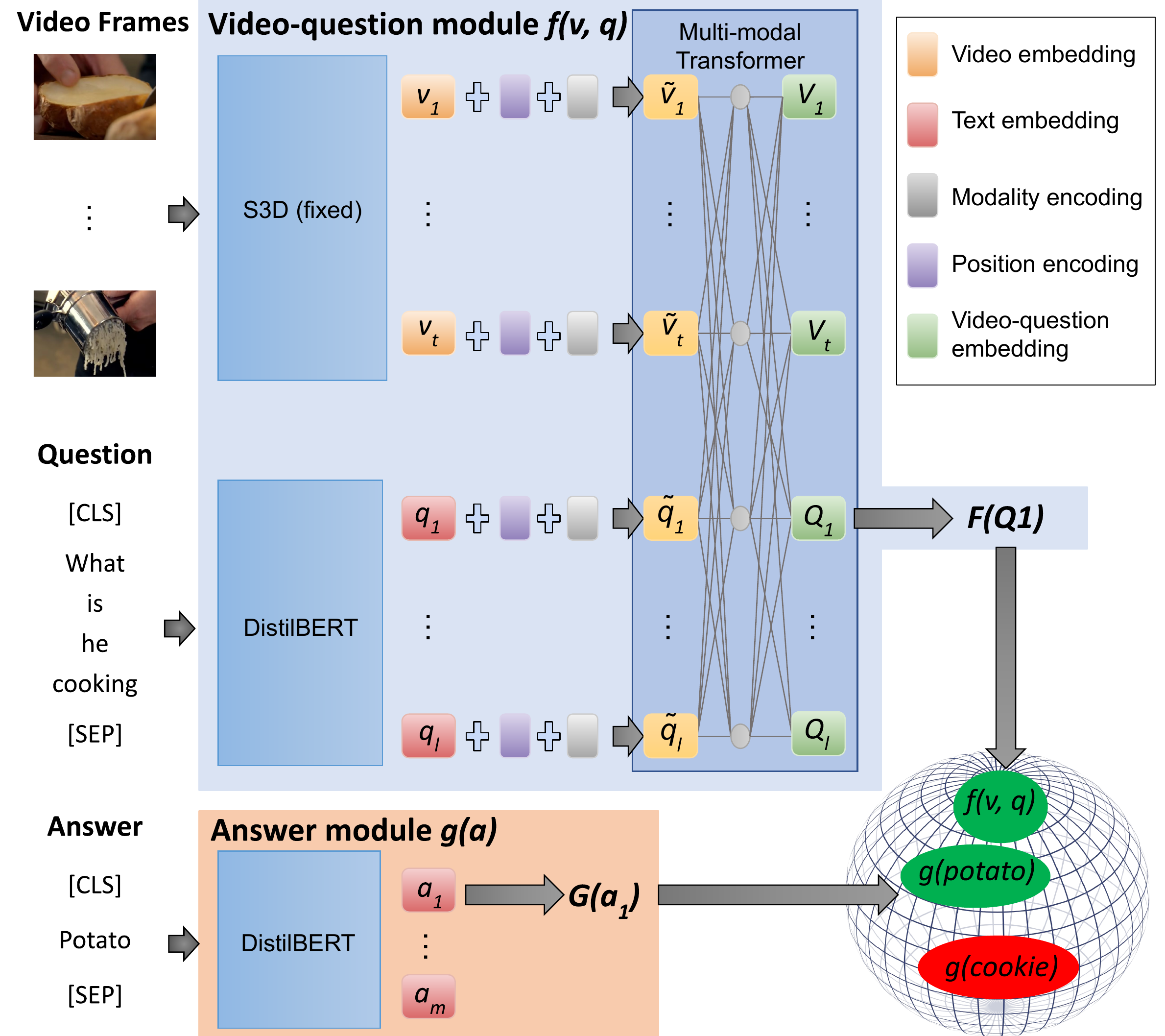}
\vspace{-0.2cm}
\caption{{\bf VideoQA architecture overview.} Our model is composed of a video-question module $f$ based on a multi-modal transformer (top) and an answer module $g$ based on DistilBERT \cite{sanh2019distilbert} encoder (bottom).}
\vspace{-0.5cm}
\label{fig:archappendix}
\end{figure*}


\section{VideoQA architecture}\label{sec:mmt}
Our architecture, shown in Figure \ref{fig:archappendix}, has two main modules: (i) a video-question multi-modal transformer (top) and (ii) an answer transformer (bottom). Details are given next, and further implementation details are given in Section \ref{sec:detailsbis}.

\begin{figure}[t]
\centering
\begin{subfigure}{.5\textwidth}
\includegraphics[width=\linewidth]{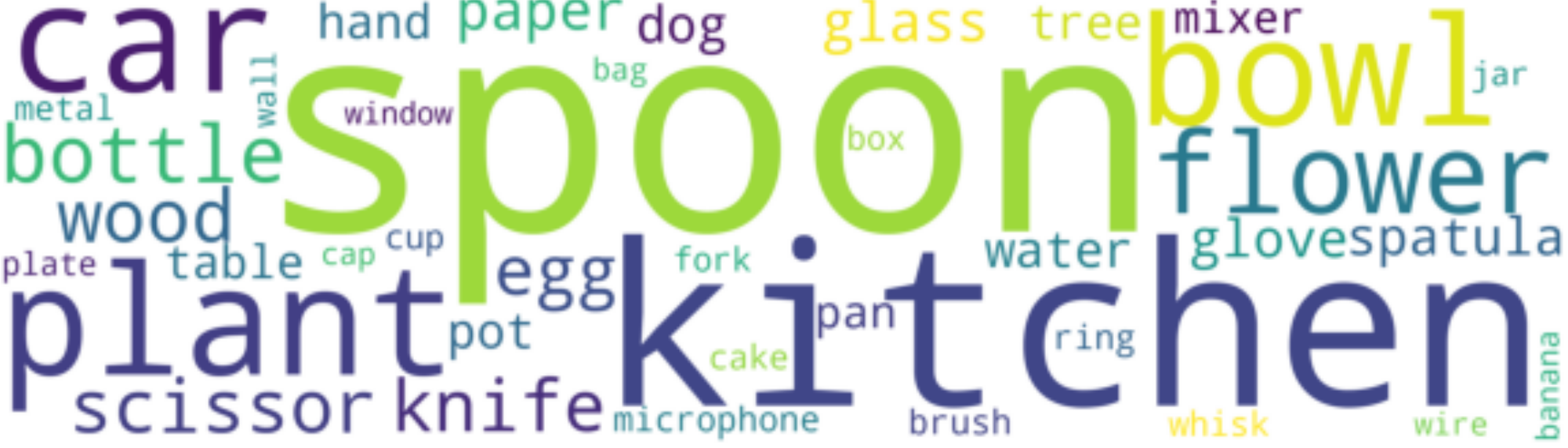}
\caption{Answers}
\end{subfigure}
\begin{subfigure}{.5\textwidth}
\includegraphics[width=\linewidth]{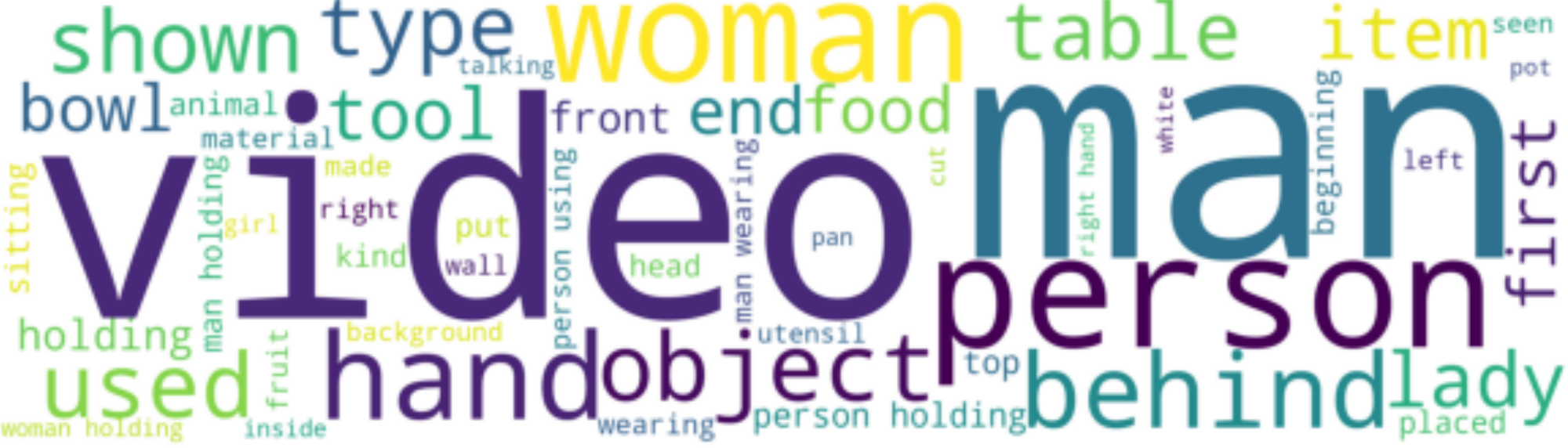}
\caption{Questions}
\end{subfigure}
\vspace{-0.31cm}
\caption{Word clouds for our \smalldatasetname{} dataset illustrate a vocabulary related to the domains of cooking, hand crafting, or gardening. The frequent  occurrence of location and time-specific words (\textit{behind}, \textit{front}, \textit{right}, \textit{left}, \textit{first}, \textit{end}, \textit{beginning}) indicate the presence of the spatial and temporal context within \smalldatasetname{} questions.}
\vspace{-0.6cm}
\label{fig:words}
\end{figure}

\begin{figure}[t]
\centering
\begin{subfigure}{.5\textwidth}
\includegraphics[width=\linewidth]{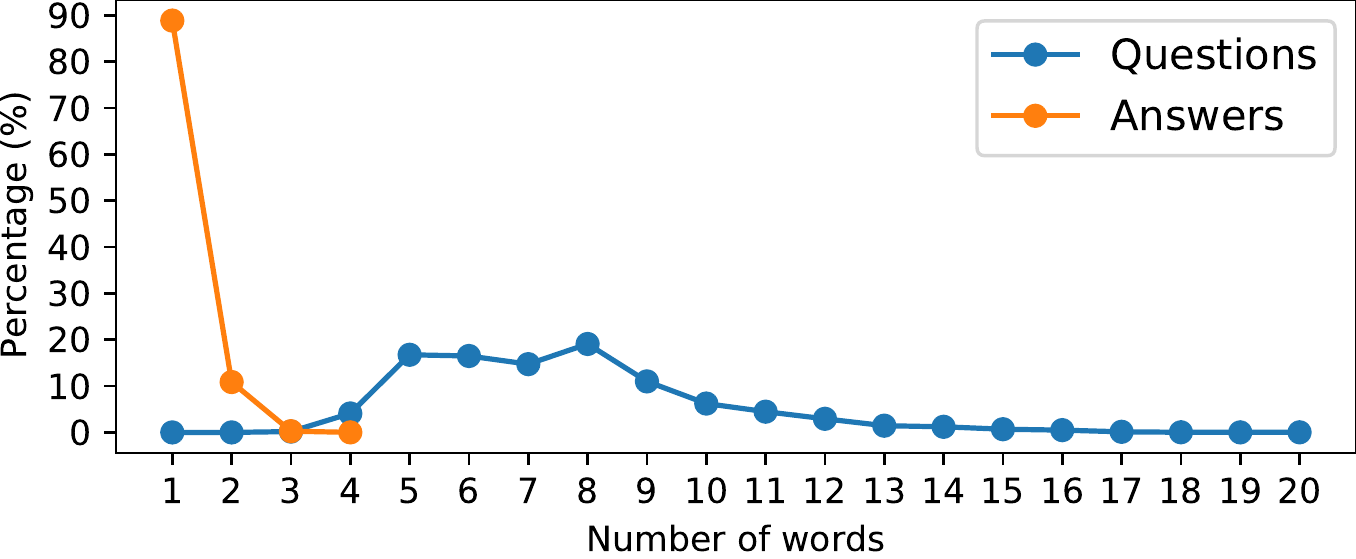}
\caption{Question and answer length}
\end{subfigure}
\begin{subfigure}{.5\textwidth}
\includegraphics[width=\linewidth]{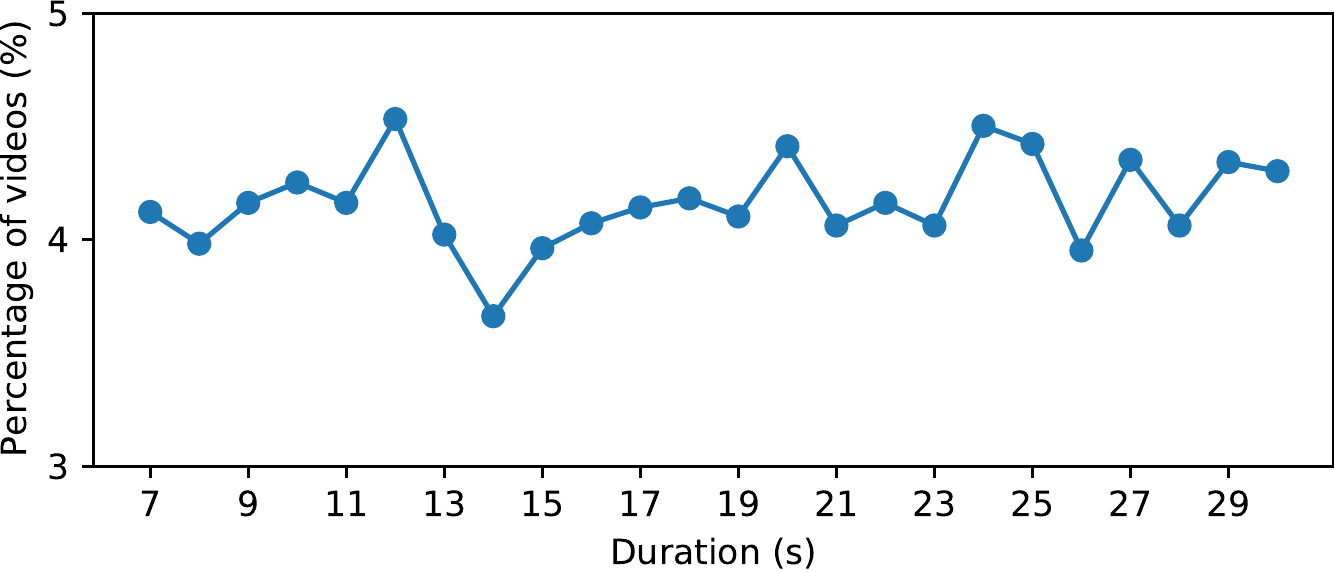}
\caption{Clip duration}
\end{subfigure}
\begin{subfigure}{.5\textwidth}
\includegraphics[width=\linewidth]{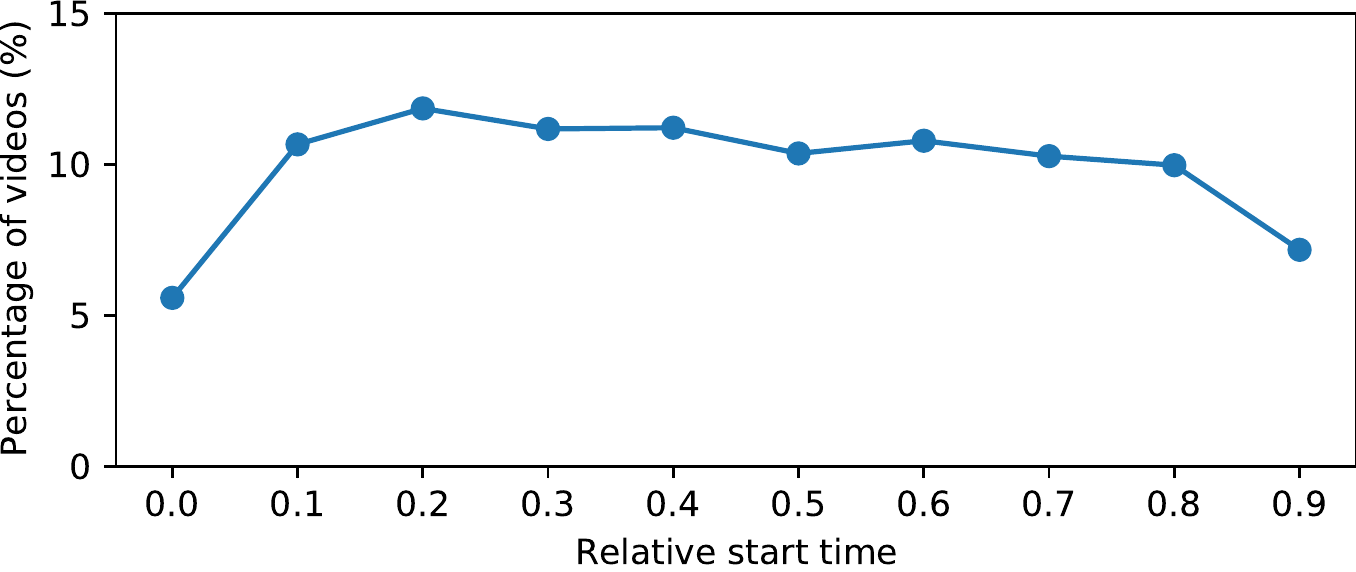}
\caption{Clip start time in the original video}
\end{subfigure}
\vspace{-.3cm}
\caption{{\bf Statistics of the \smalldatasetname{} dataset.} (a)~Distribution of length of questions and answers. (b)~Distribution of video clip duration in seconds. (c)~Distribution of video clip relative start time in the original video.}
\label{fig:length}
\vspace{-.6cm}
\end{figure}

\noindent \textbf{Video-question multi-modal transformer.} The input video representation, obtained from a fixed S3D model~\cite{xie2018rethinking}, is composed of $t$ features denoted  $v = [v_1, ..., v_t] \in \nbR^{d_v \times t}$ where $d_v$ is the dimension of the video features, and $t$ is the number of extracted features, one per second. 
The contextualized representation of the question, provided by the DistilBERT model~\cite{sanh2019distilbert}, is composed of $l$ token embeddings denoted as $q = [q_1, ..., q_{l}] \in \nbR^{d_q \times l}$ where $d_q$ is the dimension of the DistilBERT embedding and $l$ is the number of tokens in the question. 
The inputs to our video-question multi-modal transformer are then defined as a concatenation of question token embeddings and video features
\begin{equation}
\begin{split}
u(v, q) &= \left[\overset{\sim}{q}_1, ..., \overset{\sim}{q}_l, \overset{\sim}{v}_1, ..., \overset{\sim}{v}_t\right] \in \nbR^{d \times (l+t)},
\end{split}
\end{equation} 
where 
\begin{equation}\overset{\sim}{q}_s = dp\left(\sigma\left(W_q q_s + b_q\right) + pos_s + mod_q\right),\end{equation} 
and 
\begin{equation}\overset{\sim}{v}_s = dp(\sigma(W_v v_s + b_v) + pos_s + mod_v),\end{equation}
where $W_q \in \nbR^{d_q \times d}$, $b_q \in \nbR^{d}$, $W_v \in \nbR^{d_v \times d}$, $b_v \in \nbR^{d}$ and learnable parameters, $mod_q \in \nbR^{d}$ and $mod_v \in \nbR^{d}$ are learnt modality encodings for video and question, respectively, and $[pos_1, ..., pos_{l+t}] \in \nbR^{d \times (l+t)}$ are fixed sinusoidal positional encodings. $\sigma$ is a Gaussian Error Linear Unit~\cite{hendrycks2016gaussian} followed by a Layer Normalization~\cite{ba2016layer} and $dp$ refers to Dropout~\cite{srivastava2014dropout}.

\begin{figure*}[t]
\centering
\begin{subfigure}{0.95\textwidth}
\includegraphics[width=\linewidth]{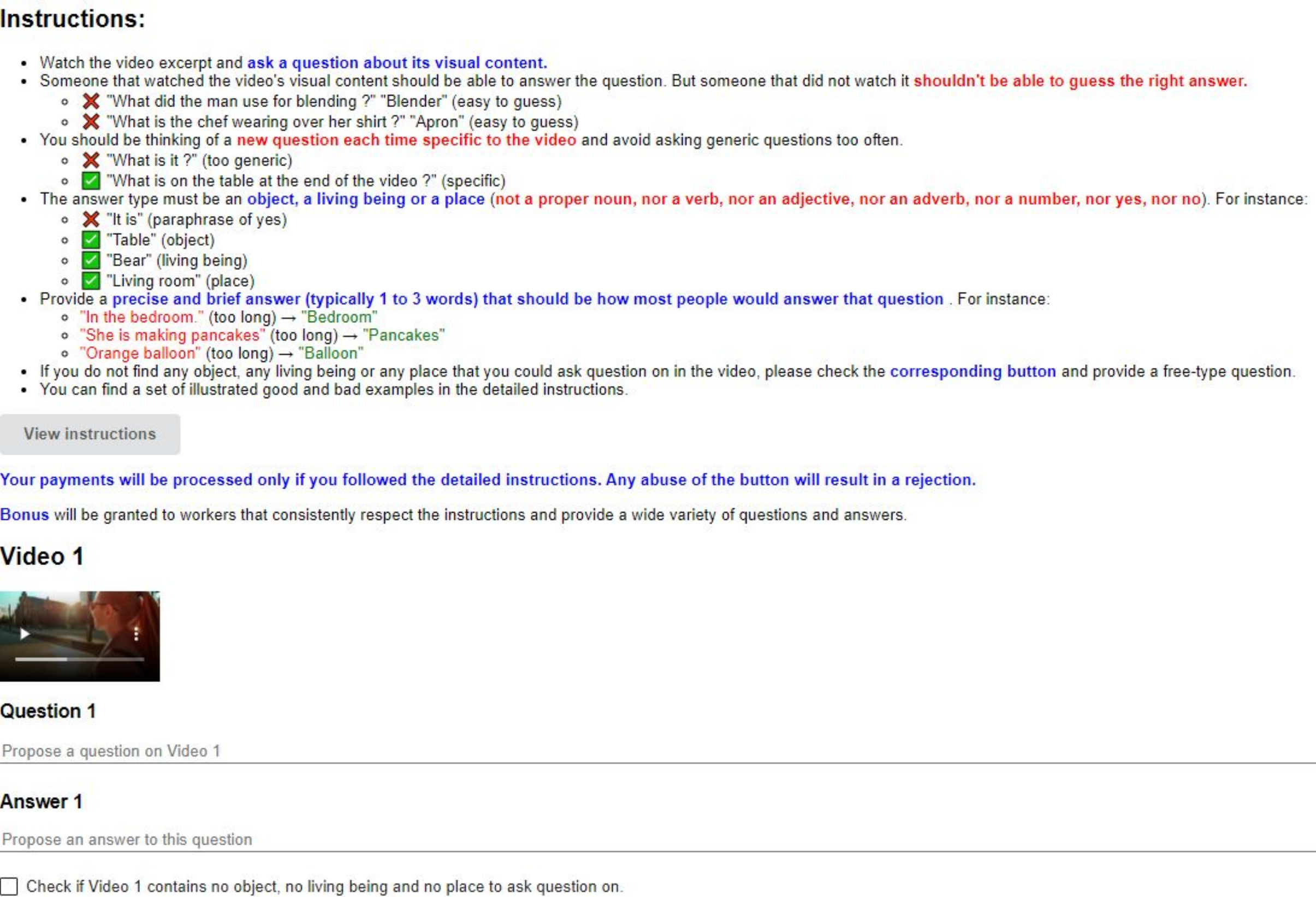}
\caption{Collection interface for questions. Note that the answer provided by the question annotator is only used to ensure that the provided question follows the given instructions, but is not included in \smalldatasetname{}. Answers are collected separately, see Figure \ref{fig:acollec}.}
\label{fig:qcollec}
\end{subfigure}
\begin{subfigure}{0.95\textwidth}
\includegraphics[width=\linewidth]{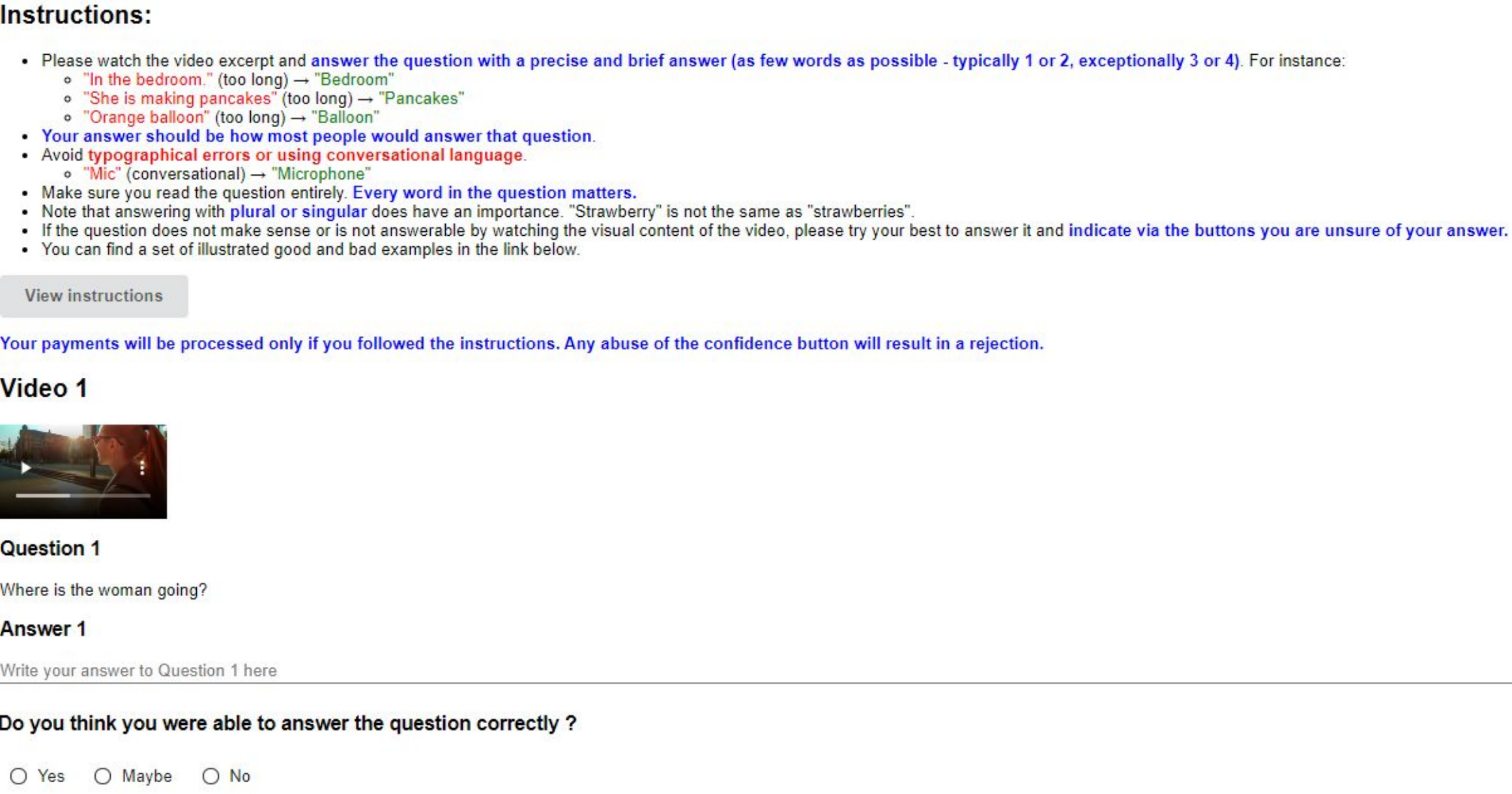}
\caption{Collection interface for answers. Five different answer annotators provide an answer annotation for each collected question.}
\label{fig:acollec}
\end{subfigure}
\vspace{-0.2cm}
\caption{Amazon Mechanical Turk interfaces for collecting questions (Figure \ref{fig:qcollec}) and answers (Figure \ref{fig:acollec}) for the \smalldatasetname{} dataset. For readability, the videos shown in these Figures are shrinked, and only one annotation example is shown.}
\vspace{-0.5cm}
\label{fig:collec}
\end{figure*}

The multi-modal transformer is a transformer with $N$ layers, $h$ heads, dropout probability $p_d$, and hidden dimension $d_h$. The outputs of the multi-modal transformer $[Q_1, ... Q_l, V_1 ... V_t] \in \nbR^{d \times (l+t)}$ are contextualized representations over tokens in the question and temporal video representations. Finally, the fused video-question embedding $f(v,q)$ is obtained as 
\begin{equation}
F(Q_1) = W_{vq} dp(Q_1) + b_{vq},
\end{equation}
where $W_{vq} \in \nbR^{d \times d}$, $b_{vq} \in \nbR^{d}$ are learnable parameters and $Q_1$ is the multi-modal contextualized embedding of the [CLS] token in the question, as shown in Figure~\ref{fig:archappendix}.

\noindent \textbf{Answer transformer.} The contextualized representation of the answer, provided by the DistilBERT model~\cite{sanh2019distilbert}, is composed of $m$ token embeddings denoted as $a = [a_1, ..., a_{m}] \in \nbR^{d_a \times m}$ where $d_a$ is the dimension of the DistilBERT embedding and $m$ is the number of tokens in the answer. 
Our answer embedding $g(a)$ is then obtained as
\begin{equation}
G(a_1) = W_a a_1 + b_a,
\end{equation}
where $W_{a} \in \nbR^{d_a \times d}$, $b_{a} \in \nbR^{d}$ are learnable parameters and $a_1$ is the contextualized embedding of the [CLS] token in the answer, as shown in Figure~\ref{fig:archappendix}.


\section{Details of the \smalldatasetname{} dataset}
\label{sec:ivqadata}

\subsection{Data Collection}
The Amazon Mechanical Turk interfaces used for collecting the question and answer annotations, are shown in Figure \ref{fig:collec}.  
An emphasis was placed on collecting visually grounded questions about objects and scenes that could not be easily guessed without watching the video, and collecting short answers in order to maximize the chance for consensus between annotators, \ie, having multiple annotators giving exactly the same answer.

\begin{table*}[t]
\setlength\tabcolsep{4.5pt}
\begin{center}
\resizebox{0.95\linewidth}{!}{	
\begin{tabular}{c|cccc|cccc}
Pretraining Data & \multicolumn{4}{c}{Zero-shot} & \multicolumn{4}{c}{Finetune} \\
& \smalldatasetname{} & \small{MSRVTT-QA} & \small{ActivityNet-QA} & How2QA 
& \smalldatasetname{} & \small{MSRVTT-QA} & \small{ActivityNet-QA} & How2QA \\ 
\hline
$\emptyset$ & --- & --- & --- & ---
& 23.0 & 39.6 & 36.8 & 80.8 \\
MSRVTT-QA 
& 8.6 & --- & 1.7 & 42.5
& 25.2 & --- & 37.5 & 80.0 \\
ActivityNet-QA & 5.5 & 2.7 & --- & 40.8
& 24.0 & 39.9 & --- & 80.7 \\
\hline
\datasetname{} & \textbf{12.2} & \textbf{2.9} & \textbf{12.2} & \textbf{51.1}
& \textbf{35.4} & \textbf{41.5} & \textbf{38.9} & \textbf{84.4} \\
\end{tabular}
}
\vspace{-0.3cm}
\caption{\small Comparison of our training on \datasetname{} with cross-dataset transfer using the previously largest open-ended VideoQA dataset (MSRVTT-QA) and the largest manually annotated open-ended VideoQA dataset (ActivityNet-QA).}
\vspace{-0.5cm}
\label{table:transfer}
\end{center}
\end{table*}

\begin{table*}[t]
\centering
\setlength\tabcolsep{7pt}
\resizebox{0.95\linewidth}{!}{
\begin{tabular}{cc|cccc|cccc|cccc}
Pretraining Data & Finetuning
& \multicolumn{4}{c}{MSRVTT-QA} 
& \multicolumn{4}{c}{MSVD-QA} 
& \multicolumn{4}{c}{ActivityNet-QA} \\
& & Q1 & Q2 & Q3 & Q4 & Q1 & Q2 & Q3 & Q4 & Q1 & Q2 & Q3 & Q4 \\
\hline
& \cmark & 
\textbf{68.4} & 44.1 & 32.9 & 8.1 &
71.2 & 53.7 & 28.9 & 8.8 &
65.6 & 49.0 & 25.7 & 3.9 \\
HowTo100M & \cmark & 
65.2 & 46.4 & 34.9 & 10.6 &
\textbf{74.8} & 58.8 & 30.6 & 10.5 &
\textbf{67.5} & \textbf{53.3} & 25.9 & 4.1 \\
\datasetname{} & \xmark & 
0.2 & 6.4 & 2.4 & 3.0 &
9.3 & 9.0 & 6.9 & 4.8 &
36.3 & 5.7 & 3.7 & 1.5 \\
\datasetname{} & \cmark & 
66.9 & \textbf{46.9} & \textbf{36.0} & \textbf{11.5} &
74.7 & \textbf{59.0} & \textbf{35.0} & \textbf{14.1} &
66.3 & 53.0 & \textbf{28.0} & \textbf{5.0} \\
\end{tabular}
}
\vspace{-0.3cm}
\caption{\small Results of our \textit{\vqat{}} model with different training strategies, on subsets of MSRVTT-QA, MSVD-QA and ActivityNet-QA, corresponding to four quartiles with Q1 and Q4 corresponding to samples with the most frequent and the least frequent answers, respectively.}
\vspace{-0.5cm}
\label{table:rareall}
\end{table*}

\subsection{Statistical Analysis}
Word clouds for questions and answers in \smalldatasetname{}, shown in Figure~\ref{fig:words}, demonstrate the relation of \smalldatasetname{} to the domains of cooking, hand crafting and gardening.
These word clouds also indicate that questions in \smalldatasetname{} often require spatial reasoning (\textit{behind}, \textit{front}, \textit{right}, \textit{left}) and temporal understanding (\textit{first}, \textit{end}, \textit{left}, \textit{beginning}) of the video. 
The most frequent answer (\textit{spoon}) in \smalldatasetname{} corresponds to 2\% of all answers in the dataset. 
In contrast, the most frequent answers in other VideoQA datasets account for more than 9\% of all answers in these datasets (we have verified this for MSRVTT-QA, MSVD-QA and ActivityNet-QA). 
As a consequence, the \textit{most frequent answer baseline} is significantly lower for our \smalldatasetname{} dataset compared to other VideoQA datasets.
Figure \ref{fig:length} shows the distributions of question length, answer length, clip duration and clip relative start time in the original video. 
Clip duration and start time distributions are almost uniform because we randomly sampled them to obtain the clips, which results in a high video content diversity.
Answers are in great majority one or two words as a result of our collection procedure.

We observe that 27.0\% of questions lead to a perfect consensus among the five answer annotators, 48.4\% of questions lead to a consensus among at least four annotators, and 77.3\% lead to a consensus among at least three annotators, while only six questions do not lead to a consensus between at least two annotators, justifying the defined accuracy metric. 
Additionally, 27.5\% of questions have two different answers that had a consensus between at least two annotators. 

\section{Additional experimental details}\label{sec:detailsbis}
\noindent \textbf{VideoQA generation.} The input sequence to the answer extractor and question generation transformers are truncated and padded up to a maximum of 32 tokens. 
The question decoding is done with the beam search keeping track of the 4 most probable states at each level of the search tree. 
We have used the original captions (including stop words) from the HowTo100M dataset~\cite{miech19howto100m} and removed word repetitions from adjacent clips.

\noindent \textbf{VideoQA model.} We use the following hyperparameters: $l=20$, $t=20$, $m=10$, $d=512$, $d_h=2048$, $N=2$, $H=8$, $p_d=0.1$, $d_q=d_a=768$, $d_v=1024$.
The video features are sampled at equally spaced timestamps, and padded to length $t$. 
Sequences of question and answer tokens are truncated and padded to length $l$ and $m$, respectively. 
Attention is computed only on non-padded sequential video and question features. 

\noindent \textbf{VideoQA datasets.} For MSRVTT-QA and MSVD-QA, we follow~\cite{le2020hierarchical} and use a vocabulary made of the top $4000$ training answers for MSRVTT-QA, and all $1852$ training answers for MSVD-QA. 
For our \smalldatasetname{} dataset and ActivityNet-QA, we consider all answers that appear at least twice in the training set, resulting in $2348$ answers for \smalldatasetname{} and $1654$ answers for ActivityNet-QA.

\begin{table*}[t]
\setlength\tabcolsep{3pt}
\begin{center}
\resizebox{.9\linewidth}{!}{	
\begin{tabular}{lc|cccccc|cccccc}
Pretraining Data & Finetuning & \multicolumn{6}{c}{MSRVTT-QA} & \multicolumn{6}{c}{MSVD-QA} 
\\ 
& & What & Who & Number & Color & When & Where
& What & Who & Number & Color & When & Where \\ \hline
& \cmark & 33.4 & 49.8 & 83.1 & 50.5 & 78.5 & 40.2 
& 31.5 & 54.9 & \textbf{82.7} & 50.0 & 74.1 & 46.4 \\
HowTo100M & \cmark & 34.3 & 50.2 & 82.7 & \textbf{51.8} & 80.0 & 41.5 
& 34.3 & 58.6 & 82.4 & \textbf{62.5} & \textbf{77.6} & \textbf{50.0} \\
\datasetname{} & \xmark & 1.8 & 0.7 & 66.3 & 0.6 & 0.6 & 4.5
& 7.8 & 1.7 & 74.3 & 18.8 & 3.5 & 0.0 \\
\datasetname{} & \cmark & \textbf{35.5} & \textbf{51.1} & \textbf{83.3} & 49.2 & \textbf{81.0} & \textbf{43.5} 
& \textbf{37.9} & \textbf{58.0} & 80.8 & \textbf{62.5} & \textbf{77.6} & 46.4 \\
\end{tabular}
}
\vspace{-0.3cm}
\caption{\small Effect of our pretraining per question type on MSRVTT-QA and MSVD-QA.}
\label{table:qtype}
\end{center}
\vspace{-0.5cm}
\end{table*}

\begin{table*}[t]
\setlength\tabcolsep{4.5pt}
\begin{center}
\resizebox{.9\linewidth}{!}{	
\begin{tabular}{lc|ccccccccc}
Pretraining Data & Finetuning & Motion & Spatial & Temporal & Yes-No & Color & Object & Location & Number & Other \\ \hline
& \cmark & 23.4 & 16.1 & 3.8 & 65.6 & 31.3 & 26.4 & 33.7 & 48.0 & 33.6 \\
HowTo100M & \cmark & 26.6 & \textbf{17.7} & 3.5 & \textbf{67.5} & 32.8 & 25.3 & 34.0 & \textbf{50.5} & 35.8 \\
\datasetname{} & \xmark & 2.3 & 1.1 & 0.3 & 36.3 & 11.3 & 4.1 & 6.5 & 0.2 & 4.7 \\
\datasetname{} & \cmark & \textbf{28.0} & 17.5 & \textbf{4.9} & 66.3 & \textbf{34.3} & \textbf{26.7} & \textbf{35.8} & 50.2 & \textbf{36.8} \\
\end{tabular}
}
\vspace{-0.3cm}
\caption{\small Effect of our pretraining per question type on ActivityNet-QA.}
\label{table:qtypeact}
\end{center}
\vspace{-0.5cm}
\end{table*}

\begin{table}[t]
\begin{center}
\setlength\tabcolsep{4pt}
\resizebox{\linewidth}{!}{	
\begin{tabular}{l|ccccc}
Method & \smalldatasetname{} & \makecell{ \small{MSRVTT} \\ \small{QA}} & \makecell{\small{MSVD} \\ \small{QA}} & \makecell{\small{ActivityNet} \\ \small{QA}} & How2QA \\
\hline
\qat{}
& 14.1 & 32.8 & 32.6 & 30.4 & 76.6 \\
\vqat{} & 23.0 & 39.6 & 41.2 & 36.8 & 80.8 \\ 
\end{tabular}
}
\end{center}
\vspace{-0.5cm}
\caption{\small Comparison of \textit{\qat{}} and \textit{\vqat{}} models trained from scratch (without pretraining) on downstream datasets.}
\vspace{-0.3cm}
\label{table:bias}
\end{table} 

\noindent \textbf{Training.} We use a cosine annealing learning rate schedule with initial values of $5 \times 10^{-5}$ and $1 \times 10^{-5}$ for pretraining and finetuning, respectively. For finetuning, we use the Adam optimizer with batch size of 256 and training runs for 20 epochs. The final model is selected by the best performance on the validation set.

\noindent \textbf{Masked Language Modeling.} For the masked language modeling objective, a token is corrupted with a probability 15\%, and replaced 80\% of the time with [MASK], 10\% of the time with the same token and 10\% of the time with a randomly sampled token. 
To guess which token is masked, each sequential question output $Q_i$ of the multi-modal transformer is classified in a vocabulary of 30,522 tokens, and we use a cross-entropy loss.

\noindent \textbf{Pretraining on HowTo100M.} For video-text cross-modal matching, we sample one video negative and one text negative per (positive) video-text pair, and use a binary cross-entropy loss. 
The cross-modal matching module is used to perform zero-shot VideoQA for the variant \textit{VQA-T} trained on HowTo100M, by computing scores for $f(v,[q,a])$ for all possible answers $a$, for each video-question pair $(v,q)$.
We aggregate adjacent clips from HowTo100M to have at least 10 second clips and at least 10 narration words. 


\section{Additional experiments}\label{sec:expbis}

\subsection{Comparison to cross-dataset transfer}\label{sec:transfer}

We define cross-dataset transfer as a procedure where we pretrain our VideoQA model on a VideoQA dataset and then finetune and test it on another VideoQA dataset. 
The training follows the procedure described for finetuning in Section \ref{sec:training}. 
We report results for cross-dataset transfer in Table \ref{table:transfer}. Note that we do not use MSVD-QA as downstream dataset as its test set has been automatically generated with the same method \cite{heilman2010good} as MSRVTT-QA.
As can be observed, our approach with pretraining on \datasetname{} significantly outperforms cross-dataset transfer models using the previously largest VideoQA dataset (MSRVTT-QA), or the largest manually annotated VideoQA dataset (ActivityNet-QA), both for the zero-shot and finetuning settings, on all four downstream datasets.
We emphasize that our dataset is generated relying on text-only annotations, while MSRVTT-QA was generated using manually annotated video descriptions and ActivityNet-QA was manually collected.
These results further demonstrate the benefit of our \datasetname{} dataset.

\subsection{Results for rare answers and per question type}\label{sec:rarebis}
Results for different answers frequencies are presented for the \smalldatasetname{} dataset in Section \ref{sec:rare}. 
Here, we show results for MSRVTT-QA, MSVD-QA and ActivityNet-QA datasets in Table \ref{table:rareall}. 
As for \smalldatasetname{}, we observe that our model pretrained on our \datasetname{} dataset, after finetuning, shows the best results for quartiles corresponding to rare answers (Q3 and Q4), notably in comparison with the model trained from scratch or the model pretrained on HowTo100M.
We also find that our pretrained model, in the zero-shot setting, performs similarly across the different quartiles, with the exception of ActivityNet-QA, which includes in its most common answers \textit{yes}, \textit{no}.
Note that in order to have a consistent evaluation with other experiments, we keep the same train vocabulary at test time. 
This implies that a significant part of answers in the test set is considered wrong because the answer is not in the vocabulary. 
This represents 16\% of answers in \smalldatasetname{}, 3\% of answers in MSRVTT-QA, 6\% for MSVD-QA and 19\% for ActivityNet-QA. 
Note, however, that our joint embedding framework could allow for different vocabularies to be used at the training and test time. 

We also present results per question type for MSRVTT-QA, MSVD-QA and ActivityNet-QA in Tables \ref{table:qtype} and \ref{table:qtypeact}. Compared to the model trained from scratch or the model pretrained on HowTo100M, we observe consistent improvements for most categories.

\subsection{Comparison between \textit{\qat{}} and \textit{\vqat{}} on different datasets.}\label{sec:bias}

We show in Table~\ref{table:bias} that \textit{\qat{}} is a strong baseline compared to \textit{\vqat{}} on existing VideoQA datasets, when both are trained from scratch.
However, on \smalldatasetname{}, \textit{\vqat{}} improves more over \textit{\qat{}} than in other datasets, as measured by absolute improvement in top-1 accuracy.
This suggests that the visual modality is more important in \smalldatasetname{} than in other VideoQA datasets.  

\end{document}